\documentclass[10pt,twocolumn,letterpaper]{article}

\usepackage{geometry}
\usepackage{cvpr}
\usepackage{times}
\usepackage{epsfig}
\usepackage{graphicx}
\usepackage{amsmath}
\usepackage{amssymb}
\usepackage{helvet}  %Required
\usepackage{courier}  %Required
\usepackage{url}  %Required
\usepackage{graphicx}  %Required
\usepackage{multirow}
\usepackage{threeparttable}
\usepackage{color}
\usepackage{booktabs}
\usepackage{array}
\usepackage{bm}
\usepackage[noblocks]{authblk}
\usepackage[dvipsnames]{xcolor}

\usepackage[breaklinks=true,bookmarks=false, colorlinks, linkcolor=red]{hyperref}
\usepackage{cleveref}

\newcommand{\bw}{\bm{w}}
\newcommand{\bx}{\bm{x}}
\newcommand{\by}{\bm{y}}
\newcommand{\bz}{\bm{z}}

\newcommand{\ft}{\textcolor[rgb]{0,0.00,0.00}}
\newcommand{\ftsec}{\textcolor[rgb]{0,0.00,0.00}}

\newcommand{\wh}{\textcolor[rgb]{0,0.00,0.00}}
\newcommand{\whf}{\textcolor[rgb]{0,0.00,0.00}}
\newcommand{\whfi}{\textcolor[rgb]{0,0.00,0.00}}
\newcommand{\whsec}{\textcolor[rgb]{0,0.00,0.00}}
\newcommand{\ftth}{\textcolor[rgb]{0,0,0}}
\newcommand{\whthi}{\textcolor{Black}}
\newcommand{\whfo}{\textcolor{Black}}

\newcommand{\whfif}{\textcolor[rgb]{0,0.00,0.00}}
\newcommand{\ftfif}{\textcolor[rgb]{0,0.00,0.00}}

\newcommand{\ftfi}{\textcolor[rgb]{0,0,0}}
\newcommand{\whsix}{\textcolor[rgb]{0,0,0}}
\newcommand{\whsev}{\textcolor[rgb]{0,0,0}}
\renewcommand{\eg}{\textit{e.g.}}
\renewcommand{\ie}{\textit{i.e.}}
\cvprfinalcopy

\def\cvprPaperID{1842} % *** Enter the CVPR Paper ID here

\ifcvprfinal\fi
\begin{document}

{\onecolumn

\noindent \vspace{1cm}

\noindent \textbf{\huge{Learning to Detect Important People in Unlabelled Images for Semi-supervised Important People Detection}}

\vspace{2cm}

\noindent {\LARGE{Fa-Ting Hong, Wei-Hong Li, Wei-Shi Zheng}}

%\Large
\vspace{2cm}

% \noindent Code is available at: \\
% \ \ \ \ \ \ \ \ \ \ \ \ \url{https://weihonglee.github.io/Projects/POINT/POINT.htm}

% \vspace{1cm}

\noindent For reference of this work, please cite:

\vspace{1cm}
\noindent Fa-Ting Hong, Wei-Hong Li and Wei-Shi Zheng.
``Learning to Detect Important People in Unlabelled Images for Semi-supervised Important People Detection \emph{Proceedings of the IEEE International Conference on Computer Vision and Pattern Recognition.} 2020.

\vspace{1cm}

\noindent Bib:\\
\noindent
@inproceedings\{hong2020semi,\\
\ \ \   title=\{Learning to Detect Important People in Unlabelled Images for Semi-supervised Important People Detection\},\\
\ \ \  author=\{Hong, Fa-Ting and Li, Wei-Hong and Zheng, Wei-Shi\},\\
\ \ \  booktitle=\{Proceedings of the IEEE International Conference on Computer Vision and Pattern Recognition\},\\
\ \ \  year=\{2020\}\\
\}
}

%\clearpage
%
%\newpage
\restoregeometry

%%%%%%%%% TITLE
\title{Learning to Detect Important People in Unlabelled Images for \\Semi-supervised Important People Detection}
% \title{Learning to Rank People in Unlabeled Image for Important People Detection}

\author[ ]{\vspace{-0.6cm}Fa-Ting Hong$^{1*}$, Wei-Hong Li$^{2}$\thanks{Equal contribution.} and Wei-Shi Zheng$^{1,3}$\thanks{Corresponding author}\vspace{-0.3cm}}
% \author[ ]{Wei-Hong Li$^{2}$\thanks{Equal contribution.}}
% \author[ ]{Wei-Shi Zheng$^{1,3}$\thanks{Corresponding author}\vspace{-0.3cm}}
% \affil[ ]{$^{1}$\small School of Electronics and Information Technology, Sun Yat-sen University, China}
\affil[ ]{\small$^{1}$ School of Data and Computer Science, Sun Yat-sen University, China}
\affil[ ]{\small$^{2}$ VICO Group, School of Informatics, University of Edinburgh, United Kingdom}
\affil[ ]{\small$^{3}$ Key Laboratory of Machine Intelligence and Advanced Computing, Ministry of Education, China.}
\affil[ ]{\tt\small  hongft3@mail2.sysu.edu.cn, w.h.li@ed.ac.uk, wszheng@ieee.org\vspace{-0.8cm}}

% \author{First Author\\
% Institution1\\
% Institution1 address\\
% {\tt\small firstauthor@i1.org}
% % For a paper whose authors are all at the same institution,
% % omit the following lines up until the closing ``}''.
% % Additional authors and addresses can be added with ``\and'',
% % just like the second author.
% % To save space, use either the email address or home page, not both
% \and
% Second Author\\
% Institution2\\
% First line of institution2 address\\
% {\tt\small secondauthor@i2.org}
% }

\maketitle

\begin{abstract}

\wh{Important people detection is to automatically detect the individuals who play the most important roles in a social event image, which requires the designed model to understand a high-level pattern. However, existing methods rely heavily on supervised learning using large quantities of annotated image samples, which are more costly to collect for important people detection than for individual entity recognition (\whf{\eg,} object recognition).}
To overcome this problem, we propose learning important people detection on partially annotated images.
Our approach iteratively learns to assign pseudo-labels \whf{to} individuals in un-annotated \ftfi{images} and learns to update the important people detection model based on data with \whf{both} labels and pseudo-labels.
To alleviate the pseudo-labelling imbalance problem, we introduce a ranking strategy for pseudo-label \ftfi{estimation}, and also 
introduce two weighting strategies: one for weighting the confidence that individuals are important people to strengthen the learning on important people and the other for neglecting noisy unlabelled images (\ie, images without any important people). We have collected two large-scale datasets for evaluation. The extensive experimental results clearly confirm the efficacy of our method attained by leveraging unlabelled images for improving the performance of important people detection.
% The proposed method learn to predict labels for unlabeled data and it can estimate the probability to alleviate the effect of those images without important people and noisy images, such that those unlabeled images where there is important people can be used and benefit the training on limited labeled data.

\end{abstract}

\section{Introduction}\label{sec:intro}
% !TEX root = main.tex
\begin{figure}[t]
	\begin{center}
		\label{fig:ShowOnFirstPage}
		%		\fbox{\rule{0pt}{2in}\rule{0.9\linewidth}{0pt}}
		\includegraphics[width=0.9 \linewidth]{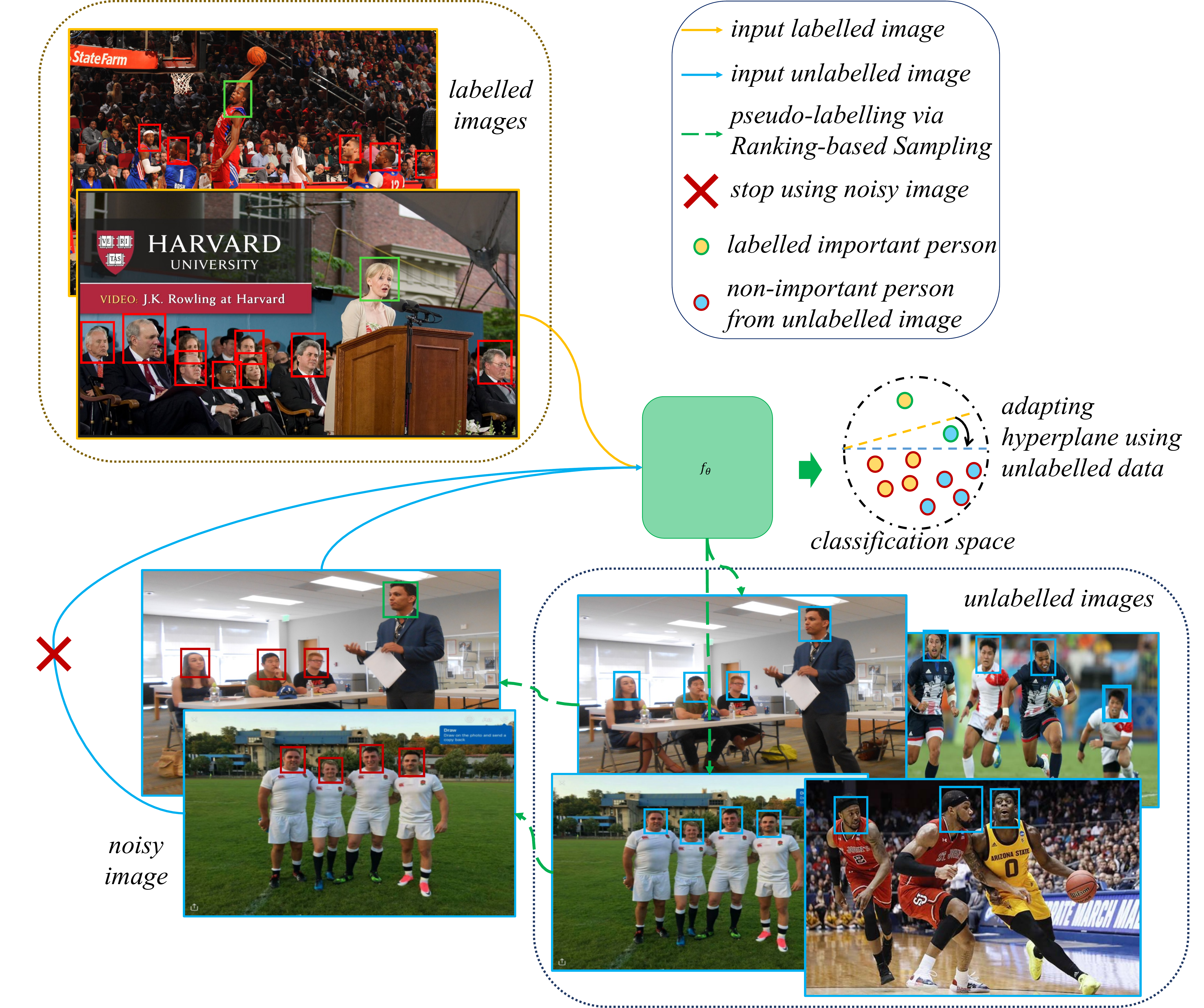}
		%		\vspace{-0.2cm}
		\centering\small\caption{
		\whsec{Collecting large quantities of labelled data for important people detection is difficult and costly. Additionally, 
% 		\whsix{since there are highly potentially important people in social event images,}
		since there are always important people in social event images,
		we design a semi-supervised method that learns to automatically select available unlabelled images as well as prevent the noisy unlabelled images  and detecting important people in unlabeled images 
		%for important people detection 
		\ft{to adapt the hyperplane of importance classification initialized by model trained with only labelled data}.}
		}   
		\label{fig:ShowOnFirstPage}
	\end{center}
	\vspace{-0.8cm}
\end{figure}

The objective of important people detection is to automatically recognize the most important people who play the most important role in a social event image. Performing this task is natural and easy for a human. This topic has attracted increasing attention, as it has a wide range of realistic applications including event detection \cite{ramanathan2016detecting}, activity/event recognition \cite{Acti_tang2017latent,ramanathan2016detecting}, image captioning \cite{VIP_solomon2015vip}, an so on. 

Developing \whfi{a} model to detect important people in images \whf{remains} challenging, as it requires the model to understand a higher-level pattern (\ftsec{\eg, relation among people in an image}) in each image compared with the information needed in other vision tasks (\eg, object-level information in classification or object detection). 
Existing methods of important people detection require massive quantities of labelled data, which are difficult and very costly to collect for this task, as it requires humans to vote on the important people \cite{li2018personrank,ghosh2018role}. 
% \whsix{Since there are highly probably important people in social event images,}
Since there always are important people in social event \whf{images},
it is natural to ask if an important people detection model can be built to learn from partially annotated data, \ie, a limited quantity of labelled data with a large number of unlabelled images. The question then arises regarding the design of \whf{a} model that can learn from partially annotated data \whf{for} important people detection if \whf{we augment} the limited labelled training data with unlabelled images.

\begin{figure}[t]
	\begin{center}
		%		\fbox{\rule{0pt}{2in}\rule{0.9\linewidth}{0pt}}
		\includegraphics[width=0.9\linewidth]{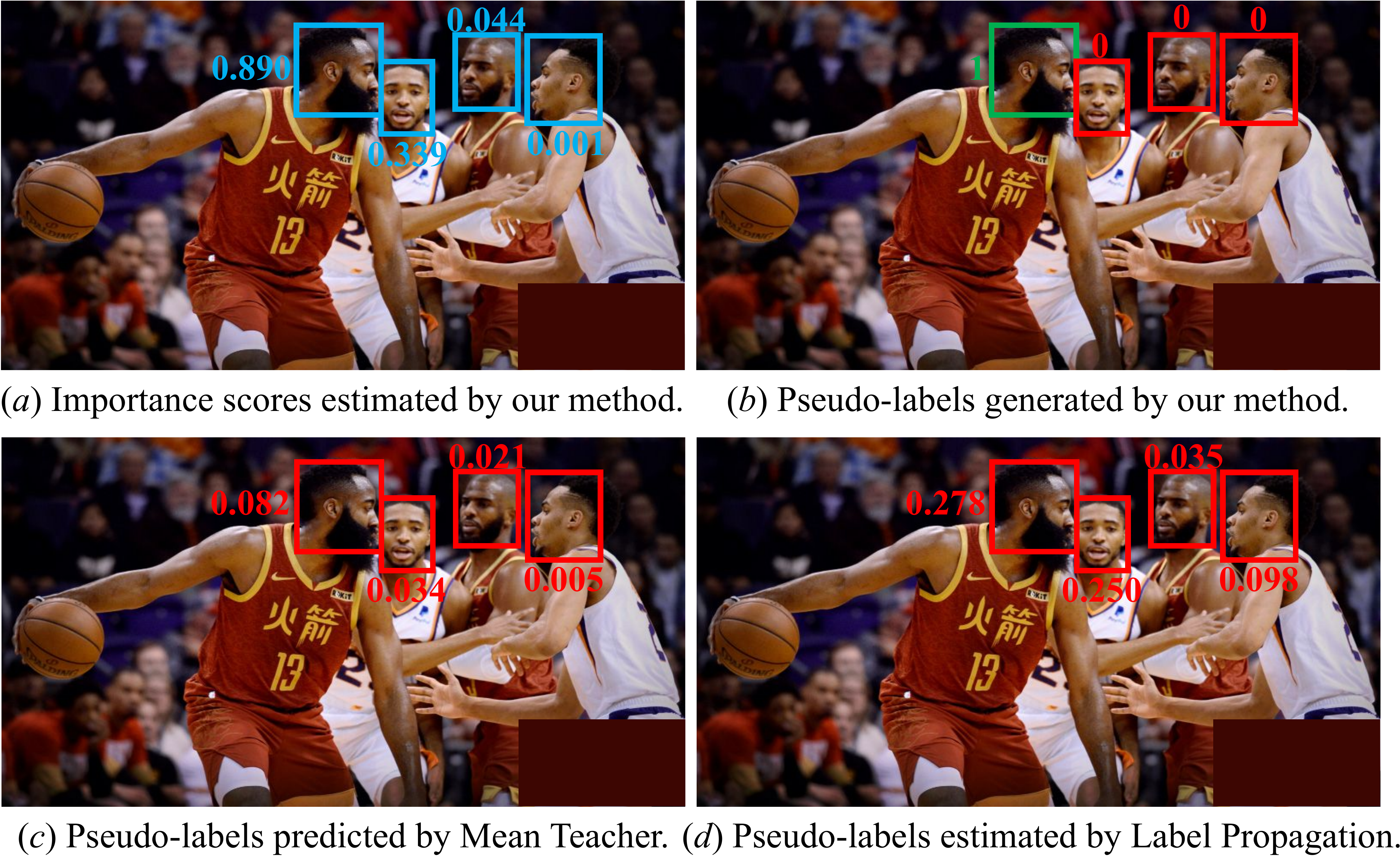}
		%		\vspace{-0.2cm}
		\centering\small\caption{
		\whfif{Examples of \whsix{our method's results and} pseudo-labels estimated by different methods during training. Blue face boxes together with numbers in Figure (a) show the importance scores generated by our method. In Figure (b), (c), (d), pseudo-labels generated by our method and related approaches are shown in terms of ``important'' category's probabilistic numbers and face boxes in different colors. Here, individuals marked with red face boxes are assigned with ``non-important' pseudo-labels and individuals marked with green face boxes are treated as ``important'' people. }
		}
		\label{fig:pseudo-labelling}
	\end{center}
	\vspace{-0.5cm}
\end{figure}

However, learning to detect important people in unlabelled images has its own challenging characteristics. First, it is not an individual entity (e.g., object) recognition \whf{task} but is rather a certain classification problem \cite{goodfellow2016deep}, relying on the \whf{relation} between people in an image. 
\whfif{Second, as the statistics \whsev{of two important people detection datasets}
% \ftfi{of the datasets used in our experiments} 
shown in Figure \ref{fig:DsStat}, most images contain more than two people, resulting in a data imbalance problem that the number of important people is always much smaller than that of non-important people;}
% Second, the number of important people and of non-important people are always imbalanced in an image, where the former is much smaller than the latter. \ftth{The statistics reported in Figure \ref{fig:DsStat} implys that most images contain more than two people and this comfirms the problem of data imbalance as the proportion of non-important people in the image is usually larger than that of important people (more details about two datasets are illustrated in Sec. \ref{sec:exp})}; 
this would yield a pseudo-labelling imbalance problem when pseudo-labels are assigned to unlabelled images, which will hamper the performance of semi-supervised learning because it is highly probable that all individuals will be regarded as \whf{``non-important''} during \whf{pseudo-labelling} \whfif{(Figure \ref{fig:pseudo-labelling}(c)(d))}. Third, not all unlabelled images contain important people; images without such people represent noisy unlabelled samples during learning.

%Moreover, Those incorrect guessed/pseudo labels of people in unlabeled images can make the imbalance problem much severer and harmper the performance. Beyond this, as those unlabeled images are clawed from the internet directly, there must be some images without any important people and these images would aggravate the imbalance problem.

To \wh{tackle the aforementioned challenges of semi-supervised important people detection}, we 
%develop a semi-supervised learning approach for important people detection to effectively leverage the information at unlabeled images to assist in the training of important people detection on limited labeled data. \wh{In general, we 
%follow the iterative bootstrapping idea from \cite{yarowsky1995unsupervised}
%where 
develop an iterative learning procedure \whsix{(Figure \ref{fig:ShowOnFirstPage})} that iteratively trains an important people detection model on data with labels or pseudo-labels and subsequently generates pseudo-labels again \whsec{of} all individuals in unlabelled images. In particular, we introduce a ranking-based sampling strategy for overcoming the imbalance problem in pseudo-label learning,
\whsix{where we rank all individuals in each unlabelled image in terms of the score of the important class}
% where we rank individuals in terms of the score of the important class \fating{of all people} in each unlabelled image 
and consider the individuals with \whsec{relatively high score (\ie, higher than a threshold)}
 % the highest scores 
% to be
\whsix{as} important people (\whf{\ie, with the pseudo-label of ``important''}) while regarding the rest as non-important individuals (\whf{\ie, with the pseudo-label of ``non-important''}). 
%\whsec{As the number of persons in an image is arbitrary, we select a fixed number of individuals from each unlabelled image for training (\ie, the important individual with the highest score and $K-1$ randomly sampled ``non-important'' people) as the fully supervised learning approach in \cite{Li_2019_CVPR}.
By using the proposed ranking-based sampling, we avoid the problem of classifying all individuals in an unlabelled image as ``non-important'' and thus the pseudo-labels of unlabelled image are more reliable \whsix{(Figure \ref{fig:pseudo-labelling}(b))}.

\begin{figure}[t]
	\begin{center}
		\label{fig:DsStat}
		%		\fbox{\rule{0pt}{2in}\rule{0.9\linewidth}{0pt}}
		\includegraphics[width=0.9 \linewidth ]{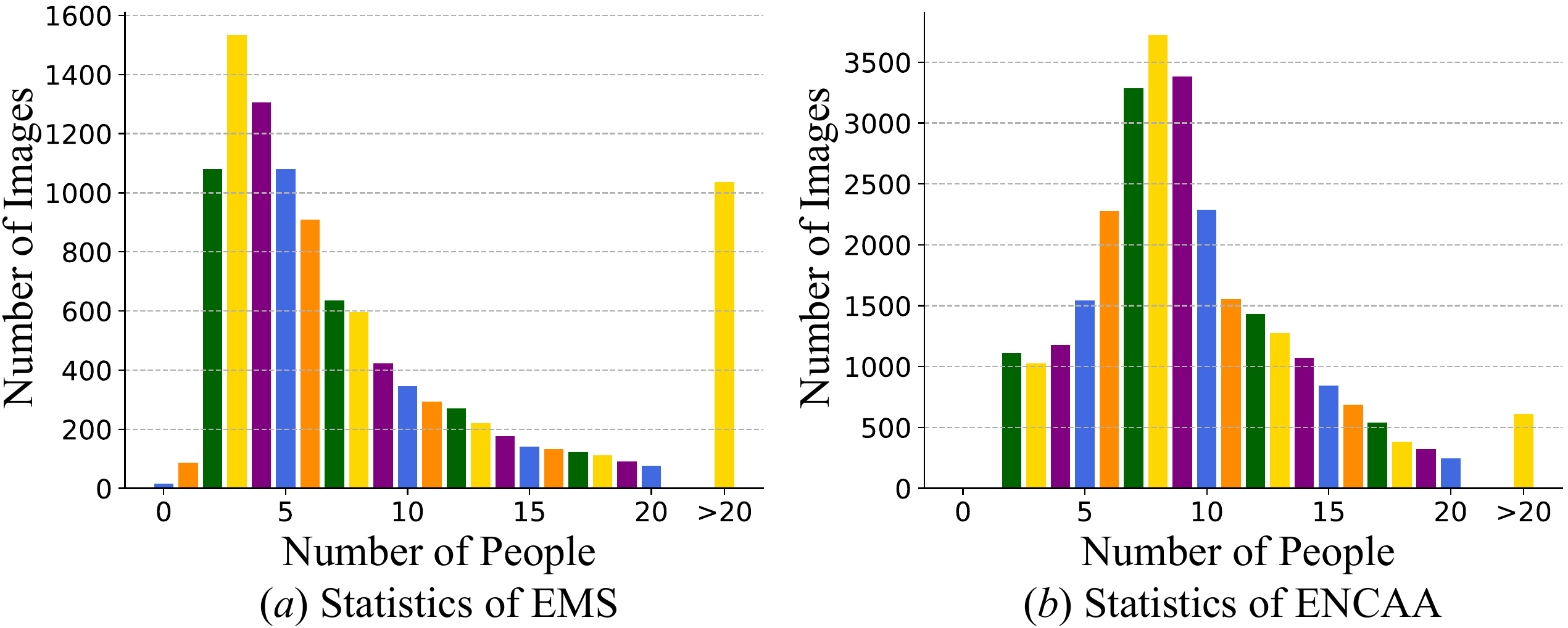}
		%		\vspace{-0.2cm}
		\centering\small\caption{\whfo{Statistics of Social Event Images in EMS and ENCAA datasets. Y-axis is the number of picture containing respective number of people (x-axis).}
		% The height of each bar represents the number of picture that contains the relative number of people.
		}   
		\label{fig:DsStat}
	\end{center}
	\vspace{-0.8cm}
\end{figure}

\ftfi{To further alleviate the pseudo-labelling imbalance problem of “non-important” data dominating the learning loss, we introduce an importance score weight to weight the confidence that individuals are important people in each unlabelled image while updating the important people detection model.}
% \whthi{caused by introducing large quantities of un-annotated images into training (\ie, the number of unlabelled images are much more than the labelled ones and the number of ``non-important'' samples are much more than ``important'' ones)}.} 
Finally, to \wh{address} the problem \whsec{caused by noisy unlabelled images (without any important people in the images), we introduce an effectiveness weight, a continuous scalar ranging from $0$ to $1$ which indicates the confidence about an unlabelled image containing important people (\ie, $0$ for no important people in the image, as opposed to $1$)} to filter out those images \ftfi{(Figure \ref{fig:ShowOnFirstPage})}. \whthi{Here, two proposed weights are estimated in a feed-forward manner at each iteration and do not require any supervisions.}

While there are \whf{no} studies on learning important people detection from partially labelled data, we contribute two large datasets called Extended-MS (EMS) and Extended-NCAA (ENCAA) for evaluation of semi-supervised important people detection by augmenting existing datasets (\ie, the MS and NCAA datasets \cite{li2018personrank}) with a large number of unlabelled images \wh{collected} from the internet.
%To the best of our knowledge, we are the first to investigate the effect of unlabeled images on important people detection and propose an effective semi-supervised learning approach \wh{for} important people detection. 
Extensive experiments verify the efficacy of our proposed method attained by \wh{enabling the labelled and unlabelled data to interact with each other and} leveraging the information of unlabelled images to assist in training of the \wh{entire} important people detection model.
% In addition, our method can be incorporated into most existing semi-supervised learning methods (\eg, mean-teacher (MT) \cite{tarvainen2017mean} and label-propagation (LP) \cite{iscen2019label}) and \wh{allows} those methods to be applied to important people detection. 
We have conducted an ablation study to investigate the effect of each component of our method (\ie, ranking-based sampling, importance score weighting and effectiveness weighting) on semi-supervised learning-based important people detection. \whsev{Additionally, the results of our method incorporating with existing semi-supervised learning approaches (\eg, mean-teacher (MT) \cite{tarvainen2017mean} and label-propagation (LP) \cite{iscen2019label}) demonstrate that our proposed method is generic and stable for semi-supervised learning-based important people detection.}

\section{Related Work}\label{sec:relwork}

\subsection{Important people/object detection}

Important people/object detection has been explored by prior work \cite{berg2012understanding,le2007finding,lee2012discovering,lee2015predicting,li2018personrank,ramanathan2016detecting,VIP_solomon2015vip,Li_2019_CVPR,ghosh2018role}, but our research is more related to the studies of important people detection \cite{ramanathan2016detecting,Li_2019_CVPR,ghosh2018role,li2018personrank,VIP_solomon2015vip}. 
To facilitate the research of important people detection, the \whsev{work \cite{li2018personrank,ghosh2018role} has} collected three small datasets, but it also indicates that it is difficult and costly to annotate a massive quantity of data for this task. These works mainly focused on developing fully supervised methods. In particular,
Ghosh et al. \cite{ghosh2018role} propose a coarse-to-fine strategy for
%where the authors first predict a group of people that contains 3 potentially important individuals and infer the most important people in that group to study the effect of group emotion on 
important people detection;
Li et al. \cite{li2018personrank} build a hybrid graph modelling the interaction among people in the image and develop a graph model called PersonRank to rank the individuals in terms of importance scores \whsec{from} the hybrid graph; In \cite{Li_2019_CVPR}, Li et al. proposed an end-to-end network called the POINT that can automatically learn the relations among individuals to encourage the network to formulate a more effective feature for important people detection.
% Editor: Please ensure that the intended meaning has been maintained in the edits of the previous sentence.

In contrast to the methods mentioned above, we mainly focus on designing a semi-supervised method to leverage information of massive unlabelled samples to assist in training a model on limited labelled data to perform important people detection.

\subsection{Learning from partially labelled data}

Learning from partially annotated data has recently become an important part of research in computer vision, as it \wh{enables} the machine learning model (deep model) to learn from a large quantity of data without costly labelling. %It has been shown that \textbf{semi-supervised learning} \cite{tarvainen2017mean,lee2013pseudo,dong2018tri,chen2018semi,iscen2019label,berthelot2019mixmatch,yarowsky1995unsupervised,oliver2018realistic,chapelle2009semi} as well as \textbf{weakly supervised learning} \cite{rocco2018end,tang2018weakly,panda2017weakly} provide an effective way to leverage information of unlabelled data to assist in training a model on limited labelled data and to boost performance, including the generalization ability \cite{chapelle2009semi}, of machine learning models. \wh{In important people detection, there are two classes (\ie, important and non-important people) in unlabelled images, which is the same as in the case of labelled data.} Thus, the setting in this study is different from that in \wh{weakly supervised} learning, and our proposed method is related to semi-supervised learning. 
Recent work \cite{tarvainen2017mean,lee2013pseudo,dong2018tri,chen2018semi,iscen2019label,berthelot2019mixmatch,grandvalet2005semi,miyato2018virtual,laine2016temporal,xie2019unsupervised} on semi-supervised learning mainly follows the well-known iterative bootstrapping method introduced in \cite{yarowsky1995unsupervised} (\ie, the classifier trained on a current set of labelled samples is used to generate labels for unlabelled data in each iteration). Among these studies, Grandvalet et al. \cite{grandvalet2005semi} proposed adding a loss term to minimize the entropy of the generated labels of unlabelled data based on the cluster assumption \cite{chapelle2009semi}. According to the latter, the work in  \cite{lee2013pseudo} proposed a method called ``Pseudo Label'' that generates a label with the maximum probability for every unlabelled sample and uses it as a true label. Another well-known assumption \cite{chapelle2009semi} is about smoothness, whereby researchers base their methods on the consistency regularization strategy to enable the model to be invariant to the added noise. Miyato et al. \cite{miyato2018virtual} introduce a consistency loss based on the predictions for an unlabelled sample with and without a learned noise to encourage the model to be invariant to the learned noise. In \cite{laine2016temporal}, the authors propose the $\Pi$ model to regularize the consistency with the models of the previous iterations by using temporal ensembles, while Tarvainen et al. \cite{tarvainen2017mean} introduce a teacher network that represents the exponential average over each iteration's model (\ie, student model) to tackle the limitation of using temporal ensembles noted in \cite{laine2016temporal}. In contrast, Iscen et al. \cite{iscen2019label} proposed a method to regularize the consistency between the prediction of the unlabelled sample and the guessed label by using the label propagation technique. Following the consistency regularization strategy, recent methods MixMatch \cite{berthelot2019mixmatch} and UDA \cite{xie2019unsupervised} embed the idea of data augmentation techniques in consistency regularization, where they regularize the model to be consistent over two augmentations of the same unlabelled images. In addition to these, Li et al. \cite{li2019learning} design a meta-learning framework that learns to impute unlabelled data such that the performance on the validation data of the model trained on these imputed data can be improved.

\wh{Unlike the above methods mainly proposed for \whf{standard} image classification, in this work, we mainly focus on developing a semi-supervised approach for important people detection, where those methods are unsuitable. \whsec{In particular, the importance of a person in an image is related to that of other people in the same image}. 
% ; \fating{\ie, this is not an individual entity recognition task but is rather a special classification problem \cite{goodfellow2016deep} relying on the interaction between people in the image.}
%which is a structured classification problem \cite{goodfellow2016deep}. 
\whsec{In contrast,} current semi-supervised approaches treat all unlabelled samples \whsec{in an unlabelled image} as independent samples and ignore the relations among them. In this paper, we design a method that can automatically exploit the pattern in unlabelled images and use a limited quantity of labelled data to assist in the overall training of an important people detection model.}
\begin{figure*}[ht]
	\begin{center}
		%		\fbox{\rule{0pt}{2in}\rule{0.9\linewidth}{0pt}}
		\includegraphics[width=1.0\linewidth]{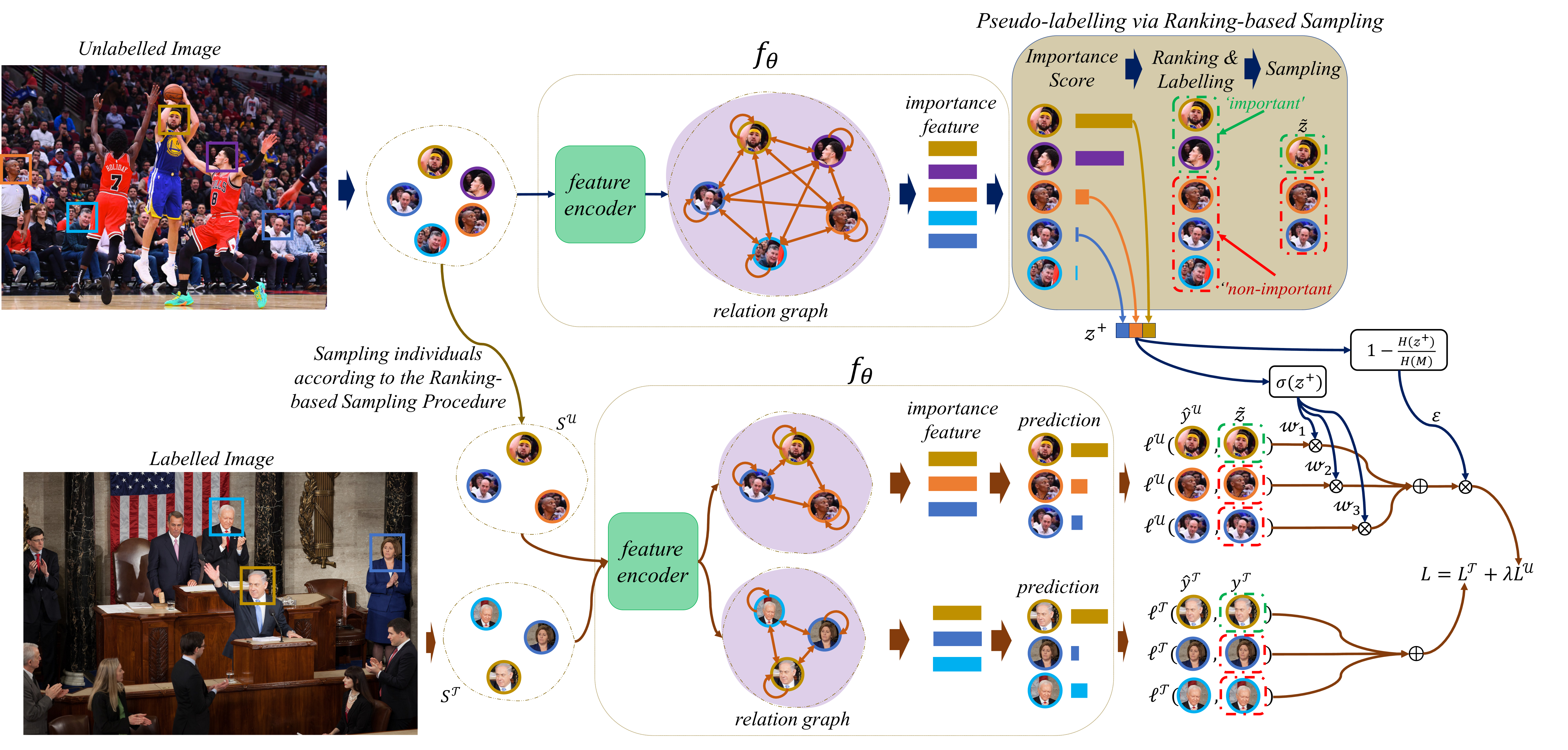}
		%		\vspace{-0.2cm}
		\centering\small\caption{
		\whsec{Illustration of our proposed framework. We feed all detected persons into $f_{\theta}$ to estimate the \whfo{pseudo-labels} by Ranking-based Sampling (RankS) according to the ranking and a threshold. We sample a fixed number of individuals in each labelled image or unlabelled image according to \whfo{labels or pseudo-labels} for training. During RankS, we also estimate the importance score weights $\bw$ as well as the effectiveness weight $\varepsilon$, which indicates the confidence that the unlabelled image features important people (\ie, $\varepsilon=1$ means that there are important people in the image, while $\varepsilon=0$ signifies the opposite), to prevent adding too many ``non-important'' individuals.}
		}
		\label{fig:FrameWork}
	\end{center}
	\vspace{-0.8cm}
\end{figure*}

\section{Methodology}\label{sec:method}

Developing a deep learning approach for important people detection requires a large quantity of labelled training data, which is difficult and costly to collect. 
% However, we observe that social event images containing multiple individuals have a high likelihood of including important people and can be used to \wh{augment} the limited labelled dataset used for training. 
To solve this problem, 
%These images can be easily collected from the internet.
%(\ie they can be collected from Internet using search queries such as ``graduation ceremony'', ``people+events'', and etc and preserving those images containing more than two people using existing people detectors).
we aim to leverage the information from unlabelled data to assist \whthi{in} training a model for important people detection on partially annotated data. An illustration of our method is shown in Figure \ref{fig:FrameWork} and is detailed in the following.

\subsection{Semi-supervised Pipeline}\label{sec:pipe}

Consider a labelled important people image dataset that contains $|\mathcal{T}|$ labelled images $\mathcal{T}=\{\mathbf{I}_i^{\mathcal{T}}\}_{i=1}^{|\mathcal{T}|}$, where for image $\mathbf{I}_i^{\mathcal{T}}=\{\bx_j^{\mathcal{T}}, y_j^{\mathcal{T}}\}_{j=1}^{N_i}$ there are $N_i$ detected persons $\bx_j^{\mathcal{T}}$ and the respective importance labels $y_j^{\mathcal{T}}$, such that $y_j^{\mathcal{T}}=0$ for the ``non-important'' class and $y_j^{\mathcal{T}}=1$ for the ``important'' class. We also have a set of unlabelled images $\mathcal{U}=\{\mathbf{I}_i^{\mathcal{U}}\}_{i=1}^{|\mathcal{U}|}$, where for each image $\mathbf{I}_i^{\mathcal{U}}$ there are $N_i$ detected individuals without any importance annotations $\mathbf{I}_i^{\mathcal{U}}=\{\bx_j^{\mathcal{U}}\}_{j=1}^{N_i}$. In this work, our goal is to design a method that can learn from partially annotated data. In other words, we aim at developing a model $\hat{\by}=f_{\theta}(\{\bx_j\}_{\bx_j \in \mathbf{I}})$ to learn from the augmented training set (\ie, $\mathcal{T} \cup \mathcal{U}$). Here, the model \whthi{$f_{\theta}$} parameterized by $\theta$ takes as input all detected individuals \ftsec{$\{\bx_j\}_{\bx_j \in \mathbf{I}}$} in a given image \ftsec{$\mathbf{I}$} and predicts the importance labels for all input persons.

\begin{figure}[t]
    \begin{center}
        
        %       \fbox{\rule{0pt}{2in}\rule{0.9\linewidth}{0pt}}
        \includegraphics[width=0.9 \linewidth]{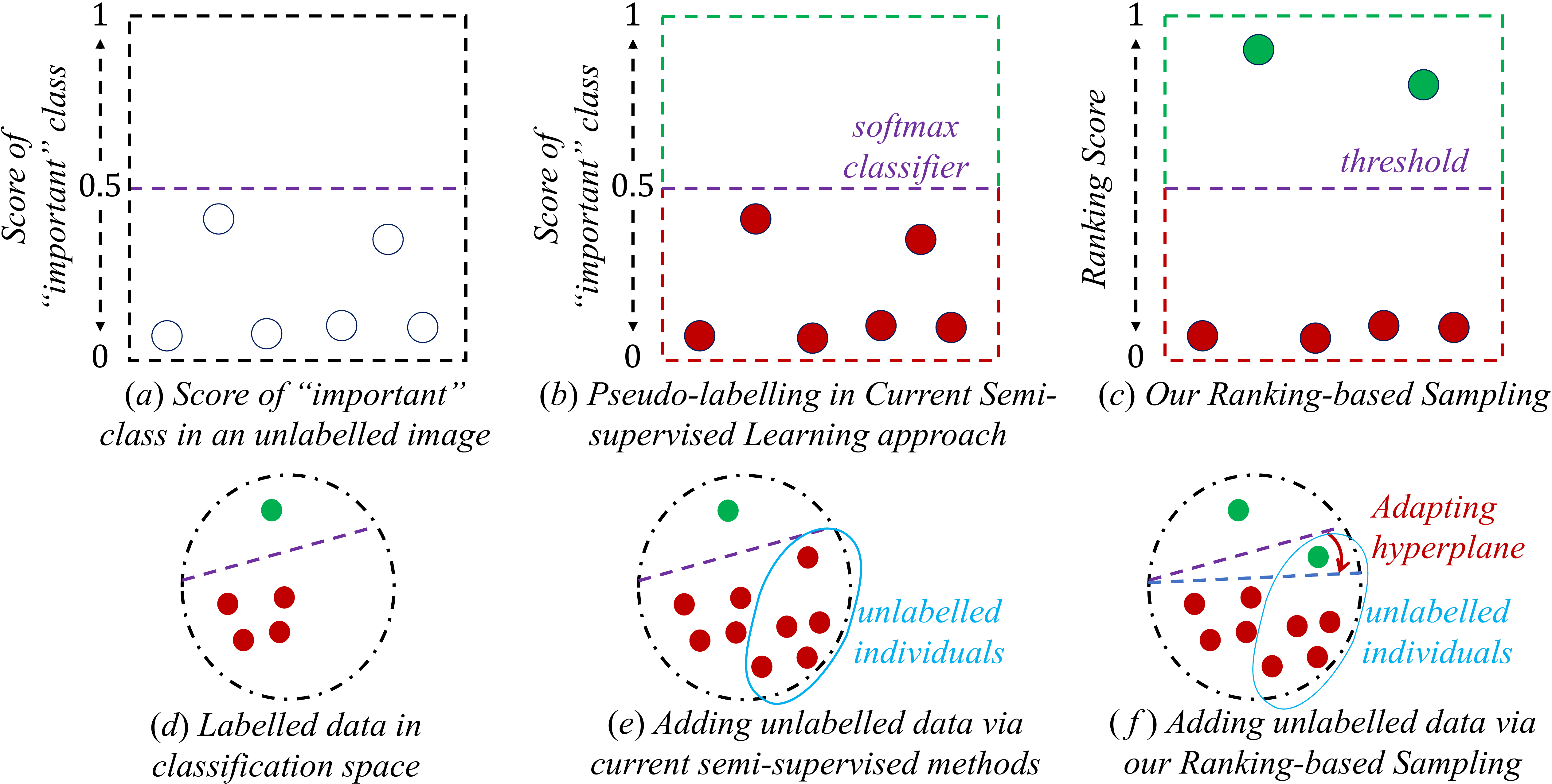}
        %       \vspace{-0.2cm}
        \centering\small\caption{\whfi{Comparison of the pseudo-labelling procedure between our method (RankS) and current semi-supervised approaches.
        \whsix{Here, dots are individuals in images: red dots indicate ``non-important'' individuals and the green ones are ``important'' people.}
        \whsec{Existing semi-supervised methods suffer the imbalanced psuedo-labelling problem that all persons in an image are always assigned as ``non-important'' class (Figure (a)-(b)) and adding those pseudo-labels doesn't help the training (Figure (d)-(e)). In contrast, our method alleviates this problem by the ranking strategy that assigns pseudo-labels according to the ranking score and threshold (Figure (c) and (f)).}
% Editor: Please ensure that the intended meaning has been maintained in the edits of the previous sentence.
    }
    \label{fig:RankS}
    \vspace{-0.8cm}
}
    \end{center}
\end{figure}

To train the model \whthi{$f_{\theta}$} on $\mathcal{T} \cup \mathcal{U}$, \whsec{we adopt \whsix{a} fully supervised model POINT \cite{Li_2019_CVPR} and formulate an iterative learning method to update this model.}
More specifically, we first train the model on one batch of labelled data (for one iteration) \whsec{to minimize the classification loss of $K$ sampled individuals ($K$ is 8 in this work as in \cite{Li_2019_CVPR}) in each labelled image. For each labelled image, we pick the ground-truth important people and randomly select $K-1$ non-important people, forming the sampled individuals set $\mathcal{S}_i^{\mathcal{T}}=\{\bx_j\}_{j=1}^{K}$ as in \cite{Li_2019_CVPR}.} \whsec{The trained model is then used to generate the pseudo-labels for $K$ sampled individuals identified in \whsec{each un-annotated image of an unlabelled images batch}
 % a batch of unlabelled images 
by $\bz=g(f_{\theta},\{\bx_j^{\mathcal{U}}\}_{\whthi{\bx_j^{\mathcal{U}}}\in\mathcal{S}_i^{\mathcal{U}}})$,
% $\bz=f_{\theta}(\{\bx_j^{\mathcal{U}}\}_{j\in\mathbf{I}^{\mathcal{U}}})$, 
where $z_j\in\bz$ is the pseudo-label of $\bx_j^{\mathcal{U}}$ estimated by pseudo-label estimation function $g(\cdot)$, and $\mathcal{S}_i^{\mathcal{U}}$ is a set of randomly sampled individuals in $\mathbf{I}_i^{\mathcal{U}}$. For \whthi{instantiating} $g(\cdot)$, 
%one can use the teacher network in MT \cite{tarvainen2017mean} or label propagation \cite{iscen2019label}, while 
we simply apply the softmax operator on the prediction \whthi{$f_{\theta}(\{\bx_j^{\mathcal{U}}\}_{\bx_j^{\mathcal{U}}\in\mathcal{S}_i^{\mathcal{U}}})$}.}
%and evaluate our method \whthi{cooperating with} MT and LP in Sec. \ref{sec:exp}.} 
\whsix{Finally, we input the unlabelled persons and their pseudo-labels into model $f_{\theta}$ for training.}

In this way, we can unify the entire training procedure as optimizing $\theta$ to minimize the following loss:
% \vspace{-0.1cm}
\begin{equation}\label{eq:loss}
\begin{footnotesize}
\begin{aligned}
    L &= L^{\mathcal{T}} + \lambda L^{\mathcal{U}}\\
      &= \frac{1}{|\mathcal{T}|K}\sum_{i=1}^{|\mathcal{T}|}\sum_{\bx_j^{\mathcal{T}}\in \mathcal{S}_i^{\mathcal{T}}}\ell^{\mathcal{T}}(\hat{y}_j^{\mathcal{T}}, y_j^{\ftsec{\mathcal{T}}}) + \frac{\lambda}{|\mathcal{U}|K}\sum_{i=1}^{|\mathcal{U}|}\sum_{\bx_j^{\mathcal{U}}\in \mathcal{S}_i^{\mathcal{U}}}\ell^{\mathcal{U}}(\hat{y}_j^{\mathcal{U}}, z_j),
\end{aligned}
\end{footnotesize}
\end{equation}
% \vspace{-0.1cm}
where $\ell^{\mathcal{T}}(\cdot)$ and $\ell^{\mathcal{U}}(\cdot)$ are a classification function (\ftfi{\ie, cross entropy}) for labelled data and a loss function for unlabelled data (\eg, mean square error) \whthi{as in \cite{tarvainen2017mean,berthelot2019mixmatch,xie2019unsupervised}}, respectively. \whthi{Additionally,
\ftsec{$\hat{y}_j^{\mathcal{T}}$} and \ftsec{$\hat{y}_j^{\mathcal{U}}$} are the predictions of a sampled individual $\bx_j^{\mathcal{T}}$ and $\bx_j^{\mathcal{U}}$ in a labelled image $\mathbf{I}_i^{\mathcal{T}}$ and an unlabelled image $\mathbf{I}_i^{\mathcal{U}}$, respectively.}
% \ftsec{The $\hat{y}_j^{\mathcal{T}}$ and $\hat{y}_j^{\mathcal{U}}$ are the prediction of the model for labelled and unlabelled data respectively and $y_j$ is the ground true of the relative people.} 
\wh{$\lambda$ is the weight of the unlabelled data loss; it is initialized to 0 and linearly increases to its maximum over a fixed number of epochs, 
%( we set the maximum as 1 and the number of epochs as 35 in this work), 
which is the well-known linear schedule \cite{berthelot2019mixmatch,xie2019unsupervised,oliver2018realistic}. Accordingly, the interaction between labelled and unlabelled data is initialized to 0 and is \wh{strengthened} gradually (\ie, during training the model is increasingly confident and the pseudo-labels of unlabelled images become more reliable).}

%\subsection{Ranking Sampling}\label{sec:RankS}

\subsection{\whsix{Pseudo-labelling} by Ranking-based Sampling}\label{sec:RankS}

An intuitive way to generate pseudo-label (either ``important'' people or ``non-important'' people) of images is to learn a classifier using limited labelled data and classify the unlabelled persons in each images. However, the number of important people and of non-important people are always imbalanced in an image, where the former is much smaller than the latter; this would yield an pseudo-labelling imbalance problem that it is highly probable that all individuals will be regarded as \whf{``non-important''}.
%when pseudo-labels are assigned to persons in unlabelled images.

To solve this problem, we design a \wh{ranking-based sampling (RankS)} strategy to predict the pseudo-labels of the individuals. Intuitively, if there are important people in an unlabelled image, \whsev{some people} must be more important than others, which forms a ranking list on the detected persons in an image as shown in Figure \ref{fig:RankS}. Therefore, we introduce the label-guessing function as
\begin{equation}
\begin{small}
    \mathcal{S}_i^{\mathcal{U}}, \tilde{\bz}=RankS(f_{\theta}, \{\bx_j^{\mathcal{U}}\}_{\whthi{\bx_j^{\mathcal{U}}}\in \mathbf{I}_i^{\mathcal{U}}}, \alpha, K),
\end{small}
\end{equation}
\whsec{where $RankS(\cdot)$ is the \wh{ranking-based sampling} procedure used to generate pseudo-labels \ftfi{$\tilde{\bz}$} based on the importance score of all individuals in $\mathbf{I}_i^{\mathcal{U}}$ by using the previous iteration's trained model $f_{\theta}$, $K$ is the number of sampled individuals in each unlabelled images as illustrated in Sec. \ref{sec:pipe} and $\alpha$ is a hyper-parameter for assigning hard pseudo-labels to $K$ sampled individuals according to the ranking.}

\whsec{To be detailed, we use the relation network in \cite{Li_2019_CVPR} as the backbone (\ie, $f_{\theta}$), which takes as input a set of people per image to build relation graph and encode features for importance classification from the relation graph.
We first apply the model trained at the previous iteration \whthi{and the softmax operator} to compute the probability of the ``important'' category as the importance score, ranking all individuals detected in the same unlabelled image in the descending order of importance score by scaling the importance score with the maximum score, resulting in a ranking score. We then assign ``important'' pseudo-label to those individuals whose ranking score is higher than a threshold $\alpha$ and the rest are assigned as ``non-important''. We pick the top-1 ``important'' individual and randomly select $K-1$ ``non-important'' people to form $\mathcal{S}_i^{\mathcal{U}}$, which is used for training coupled with the pseudo-labels of sampled individuals.} 

Therefore, replacing the pseudo-labels with those generated by RankS allows the unlabelled loss term in Eq. \ref{eq:loss} to be rewritten as
\begin{equation}\label{eq:lossU}
\begin{small}
    L^{\mathcal{U}} = \frac{1}{|\mathcal{U}|K}\sum_{i=1}^{|\mathcal{U}|}\sum_{\whsev{\bx_j^{\mathcal{U}}} \in \mathcal{S}_i^{\mathcal{U}}}\ell^{\mathcal{U}}(\hat{y}_j^{\mathcal{U}}, \tilde{z}_j).
\end{small}
\end{equation}

\whsec{In this way, we regularize the consistency between the sampled \whthi{individuals' pseudo-labels estimated}
 % \ftsec{pseudo-label labelled} 
from the full relation graph (\ie, relation graph of all individuals in $\mathbf{I}_i^{\mathcal{U}}$) and the prediction from the sub-graph (\ie, the relation graph of $\mathcal{S}_i^{\mathcal{U}}$) as illustrated in Figure \ref{fig:FrameWork}. \ft{That is, we force a constraint that the importance of the person estimated based on a subset of persons selected by our Ranking sampling should be close to the one estimated from all person detected in an image.} Thanks to the ranking and labeling in RankS, the pseudo-labels $\hat{z}$ alleviate the imbalance problem \whf{to some extent}, as this approach avoids the problem of assigning all people in an image the ``non-important'' label during pseudo-labelling (Figure \ref{fig:RankS} \whsix{and \ref{fig:pseudo-labelling}}).}

% We also verify that, the ranking sampling strategy introduced 

\subsection{Balancing Loss via Importance Score \ftsec{Weighting}}\label{sec:ISW}

Still, we treat the respective people in unlabelled images equally, while there are much more ``non-important'' samples than ``important'' samples in unlabelled images, which could make the pseudo-labelling imbalance problem remain. 

To further alleviate the pseudo-labelling imbalance problem, instead of \wh{assigning} each person in each unlabelled image \wh{the same weight} (\ie, $\frac{1}{K}$ for $K$ sampled individuals in an image), we introduce a person-specific weight $w$, \ftsec{called importance score weight (ISW),} into the unlabelled data loss term in Eq. \ref{eq:lossU} \whsix{so that}
% to alleviate the imbalance problem, so that 
% we expect 
the contribution of the ``important'' person will be \wh{strengthened}, and those of ``non-important'' ones will be \wh{weakened}.
%/***Jason: need to tell the effect of the weight. please fill...***/. 
% \vspace{-0.5cm}
For this purpose, we can rewrite Eq. \ref{eq:lossU} as
\begin{equation}\label{eq:lossISW}
\begin{small}
\begin{aligned}
    L^{\mathcal{U}}=\frac{1}{|\mathcal{U}|}\sum_{i=1}^{|\mathcal{U}|}\sum_{\whsev{\bx_j^{\mathcal{U}}} \in \mathcal{S}_i^{\mathcal{U}}}w_j\ell^{\mathcal{U}}(\hat{y}_j^{\mathcal{U}}, \tilde{z}_j),~~
    &\text{s.t.}~~\sum_{j=1}^{K}{w_j} = 1, w_{j} > 0.
\end{aligned}
\end{small}
\end{equation}

% \begin{equation}\label{eq:lossISW}
% \begin{small}
% \begin{aligned}
%     L^{\mathcal{U}}&=\frac{1}{|\mathcal{U}|}\sum_{i=1}^{|\mathcal{U}|}\sum_{\bx_j \in \mathcal{S}_i^{\mathcal{U}}}w_j\ell^{\mathcal{U}}(\hat{y}_j^{\mathcal{U}}, \tilde{z}_j)\\
%     &\text{s.t.}\quad \sum_{j=1}^{K}{w_j} = 1, w_{j} > 0.
% \end{aligned}
% \end{small}
% \end{equation}
% \vspace{-0.5cm}

To estimate the weight $w_j$ of person \whsev{$\bx_j^{\mathcal{U}}$}, we first consider the probability of the ``important'' class $z^{+}_j$ for \ftfi{$\bx_j^{\mathcal{U}}$} and treat $z^{+}_j$ as the importance score. As we mentioned in Sec. \ref{sec:RankS}, given an unlabelled image, $K$ persons are sampled, and their importance scores form an importance score vector $\bz^{+}$. We then apply the normalization function to $\bz^{+}$, which results in normalized importance score weights $\bw=\sigma(\bz^{+})=(w_1,w_2,\cdots,w_K)$, where $\sigma(\cdot)$ is a normalization function applied to $\bz^{+}$, so that the constraint in Eq. \ref{eq:lossISW} is satisfied. In this work, instead of using a hard importance score weight (\ie, \wh{$w_j \in \{0,1\}$}), we use softmax to obtain a soft importance score weight ( \ie, $w_i \in [0,1]$), so that our model has a better tolerance on the bias of computing importance scores.
\whthi{Here, we do not apply \ftsec{importance score weighting} to labelled data for couple reasons. First, the number of unlabelled images is much larger than that of labelled data and the imbalance problem has been largely mitigated by \ftsec{importance score weighting}. Second, as ``non-important'' individuals in labelled data have ground-truth annotations, we consider using this more reliable information and weakening the effect of unlabelled ``non-important'' individuals.}

\subsection{Detecting Noisy Unlabelled Images}\label{sec:EW}

Apart from the imbalance problem, it is essential that the model should \whthi{be able to} detect and neglect noisy images with no detected important people. For unlabelled images, 
%The reason is that unlabelled images were directly crawled from the internet, and 
it is not guaranteed that all images contain important people. 
%It has also been discussed in Sec. \ref{sec:ISW} that augmenting the labelled data with more ``non-important'' samples will not benefit model training or will even aggravate the imbalance problem. 
To solve this problem, we further estimate an effectiveness weight (EW) $\varepsilon$, a continuous varying value between 0 and 1, reflecting the confidence that an unlabelled image features important people (\ie, $\varepsilon=1$ means that there are important people in the image, while $\varepsilon=0$ signifies the opposite). We apply this weight into Eq. \ref{eq:lossISW} as follows:
\begin{equation}\label{eq:lossEW}
\begin{small}
\begin{aligned}
    L^{\mathcal{U}}=&\frac{1}{|\mathcal{U}|}\sum_{i=1}^{|\mathcal{U}|}\varepsilon_i\sum_{\whsev{\bx_j^{\mathcal{U}}} \in \mathcal{S}_i^{\mathcal{U}}}w_j\ell^{\mathcal{U}}(\hat{y}_j^{\mathcal{U}}, \tilde{z}_j)\\
    &\text{s.t.}\quad w_j \in \bw=\sigma(\bz^{+}), 0 \leq \varepsilon_i \leq 1,
\end{aligned}
\end{small}
\end{equation}where $\varepsilon_i$ acts as a gate to enable the model to choose or neglect the $i-th$ unlabelled image. Inspired by \cite{iscen2019label}, we consider specifying $\varepsilon$ using the entropy of the importance score $\bz^{+}$. \wh{In particular,} if there are important persons, those persons' importance scores will be high, and the other people's importance scores will remain low (\ie, the entropy will be low). In contrast, if there are no important people, the importance scores of all persons in the respective unlabelled image will be almost uniform (\ie, the entropy will be high). To constrain $\varepsilon$ between 0 and 1, we specify $\varepsilon$ as
\begin{equation}\label{eq:EW}
\begin{aligned}
\begin{small}
\varepsilon = 1 - \frac{H(\bz^{+})}{H(M)},
\end{small}
\end{aligned}
\end{equation}
where $H(\cdot)$ is the entropy function. Additionally, $M$ is a vector with the same dimension \wh{as} $\bz^{+}$, and all elements of $M$ are equal. Vector $M$ in Eq. \ref{eq:EW} \wh{simulates} the possible case of no important people, and thus, $H(M)$ is the maximum possible entropy of each unlabelled image. In this equation, if there are no important people in an unlabelled image, $H(\bz^{+})$ will be equal to $H(M)$, \wh{resulting in $\varepsilon=0$}, \ie, noisy unlabelled images will be \wh{neglected, or their effect will be weakened}.

Therefore, replacing the unlabelled data loss term in Eq. \ref{eq:loss} with Eq. \ref{eq:lossEW}, we formulate our complete method as
\begin{equation}\label{eq:lossF}
\begin{footnotesize}
\begin{aligned}
    L &= L^{\mathcal{T}} + \lambda L^{\mathcal{U}}\\
      &= \frac{1}{|\mathcal{T}|K}\sum_{i=1}^{|\mathcal{T}|}\sum_{\whsev{\bx_j^{\mathcal{T}}} \in \mathcal{S}_i^{\mathcal{T}}}\ell^{\mathcal{T}}(\hat{y}_j^{\mathcal{T}}, y_j^{\ftfi{\mathcal{T}}}) + \frac{\lambda}{|\mathcal{U}|} \sum_{i=1}^{|\mathcal{U}|}\varepsilon_i\sum_{\whsev{\bx_j^{U}} \in \mathcal{S}_i^{\mathcal{U}}}w_j\ell^{\mathcal{U}}(\hat{y}_j^{\mathcal{U}}, \tilde{z}_j)\\
    &\quad \text{s.t.}\quad w_j \in \bw=\sigma(\bz^{+}), \varepsilon_i = 1 - \frac{H(\bz^{+})}{H(M)}.
\end{aligned}
\end{footnotesize}
\end{equation}
By introducing three strategies (\ie, ranking-based sampling, importance score weighting and effectiveness weighting) to the basic semi-supervised learning pipeline, our proposed method enables the collaboration between labelled data and unlabelled data to benefit the overall model training. 

%Compared with current semi-supervised approaches, our method can use the model trained on the current iteration to more correctly detect important people in unlabelled images during training, resulting in better performance on testing.
 % be trained on labeled data and detect important people in unlabeled images for training.}
% be trained on partially annotated data effectively.

\section{Experiments}\label{sec:exp}

% !TEX root = main.tex
\wh{In this work, we conduct extensive experiments on two large datasets collected in the course of this research to investigate the effect of the use of unlabelled images on important people detection and evaluate our proposed semi-supervised important people detection method.} \ftsec{More detailed information about datasets, \ftfi{essential network (\ie, POINT)} and additional experiments results are reported and analyzed in the Supplementary Material.}

% In this work, we form two large datasets for semi-supervised important people detection and conduct experiments on both datasets to investigate the effect of unlabeled images on important people and evaluate our proposed semi-supervised important people detection method.
% In order to evaluate our proposed method exploiting unlabeled data to assist in training on limited labeled data for important people, we augment two existing datasets proposed in \cite{Li_2019_CVPR} by collecting unlabeled images online.
% \vspace{-0.05cm}

\subsection{Datasets}

\wh{Due to the lack of datasets for semi-supervised \whsev{learning-based} important people detection,} we augment two datasets, namely, the MS and NCAA datasets from \cite{li2018personrank}, by collecting a large number of unlabelled images from the internet, and forming the \ftfi{Extended}-MS (EMS) and \ftfi{Extended}-NCAA (ENCAA) datasets.
% In this work, we contribute two datasets for weakly supervised learning evaluation on important people detection. 

\noindent \textbf{The EMS Dataset} contains $10,687$ images featuring more than six types of scenes, of which $2310$ images are from the MS dataset, and $8377$ images were obtained by directly crawling the web \footnote{We collected unlabelled images from the internet by searching for various social event topics such as ``graduation ceremony''.}. 
% Editor: In the previous sentence, please consider replacing "current" with "modern" or "state-of-the-art".
For both labelled data and unlabelled data, a face detector \cite{li2019dsfd} is used to detect all possible persons, and bounding boxes are provided. 
%For labeled data, the persons who are the important people in images are annotated as $1$ and those ``non-important'' people are annotated as $0$.
Similar to \cite{Li_2019_CVPR}, the EMS dataset is split into three parts: a training set ($8607$ images, consisting of $690$ labelled samples and $8377$ unlabelled samples), a validation set ($230$ labelled samples), and a testing set ($1390$ labelled samples).

\noindent \textbf{The ENCAA Dataset}. Based on $9736$ labelled \wh{images} from the NCAA dataset, we collect $19,062$ images from the internet by extracting frames from numerous basketball videos and filtering \wh{out} images that do not feature multiple players.
% The data collected in this way contains more comprehensive information. 
Similar to the construction of the EMS dataset, we also divide the ENCAA dataset into three parts: $2825$ labelled samples picked randomly and all unlabelled samples form the training set; $941$ randomly selected labelled samples are used as a validation set; and the remaining labelled samples (\ie, $5970$ images) constitute the testing set. Each person's bounding box is generated by \wh{an} existing object detector, namely, the YOLOv3 detector \cite{redmon2018yolov3}. 
%In labeled data, the annotation (\ie ``important'' and ``non-important'') of each persons are provided while those people in unlabeled images are not annotated.

\subsection{Baselines}

In this work, we use the state-of-the-art fully supervised method as the baseline to evaluate our methods. In addition, we adapt three recent semi-supervised methods, namely, Pseudo Label (PL) \cite{lee2013pseudo}, Mean Teacher (MT) \cite{tarvainen2017mean} and Label Propagation (LP) \cite{iscen2019label}, to important people detection.

\noindent \textbf{POINT}.
We adopt the POINT \cite{Li_2019_CVPR} method, a state-of-the-art method of important people detection, as the baseline, which we train \ftsec{only} on labelled data using a fully supervised learning approach.

\noindent \textbf{Pseudo Label}
% Pseudo-labelling 
is a simple yet efficient semi-supervised learning approach for ordinary classification tasks, which chooses the class with the maximum predicted probability as the true label for each unlabelled sample. 

\noindent \textbf{Mean Teacher} maintains two models: student and teacher. Given unlabelled samples, the outputs of the teacher model are used as pseudo-labels. The consistency loss is determined over the predictions of unlabelled images \ftfi{predicted by} student model and the pseudo-labels generated by the teacher model such that the learned model can be invariant to stochastic noise between student and teacher models.

\noindent \textbf{Label Propagation} infers the pseudo-labels of unlabelled samples from the nearest neighbour graph, \whsev{which is constructed based on the embeddings of both labelled and unlabelled samples.}
% \ftfi{which is composed of the embedding of samples.}
%encoded by the network. It assigns labels to unlabelled images according to the similarity between unlabelled samples' embeddings and the labelled samples' embeddings.

% The main idea is to determine the similarity of the class label according to the similarity of embedding, making the label in constructed graphs flatter.

\subsection{Implementation Details}

% We provide training details as well as the hyper-parameters per algorithm and our approach.
% We implement all methods including all baselines and our method in PyTorch.
\whthi{We implement all methods in PyTorch.}
%Specifically, we take the code of the fully supervised approach (\ie POINT) from \cite{Li_2019_CVPR} and adapt recent semi-supervised methods, including PL, MT and LP by using the POINT as backbone for important people detection. 
%For fair comparison, we also adopt the POINT as the backbone in our method and use the commonly used SGD as the optimizer. 
For a fair comparison, we adopt POINT \ft{(we have detailed it in Supplementary Material)} as the essential network with SGD used as the optimizer in our method as well as other semi-supervised baselines (\ie, PL, MT and LP).
We run all methods for $200$ epochs and use the same hyper-parameters for all methods. \whsec{The hyper-parameter $\alpha$ is learned on the validation data and is set to 0.99 for all the experiments.} The weight decay is $0.0005$ and the momentum is \ftfi{$0.9$} in all experiments. The learning rate is initialized to \ftfi{$0.001$}, and we follow the learning rate update strategy of \cite{Li_2019_CVPR}, \ie, the learning rate is scaled by a factor of $0.5$ every $20$ epochs. We adopt the commonly used linear schedule to update weight $\lambda$, \ie, we increase $\lambda$ linearly from $0$ to its maximum (\ie, $1$) over $35$ epochs. \whfo{We follow the standard evaluation metric in}
% use the evaluation metric of 
\cite{Li_2019_CVPR}, \ie, the mean average precision is reported to measure the performance of all methods.

\subsection{Comparisons with Related Methods}

\whthi{We first compare our method with current semi-supervised learning methods adapted for important people detection and the fully supervised learning baseline.}
From Table \ref{tab:Comp}, it is worth noting that the recent semi-supervised learning approaches attain comparable results (\eg, the results of LP vs. those of POINT are $88.61$ \% vs. $88.21$ \% on the ENCAA dataset if 66 \% of labelled images are used) but sometimes underperform the fully supervised baseline (\eg, the results of LP vs. those of POINT are $86.66$ \% vs. $88.48$ \% on the EMS dataset if all labelled images are used). In contrast, our method \wh{achieves} a significant \whthi{and consistent} improvement over the baseline; \eg, After adding unlaballed images, \whthi{our method} outperforms the fully supervised baseline by $4.45$ \% and $4.15$ \% on the EMS and ENCAA datasets, respectively, in the regime with fewer \wh{labels} (33 \%). These results of PL, LP and MT clearly demonstrate that treating each person independently are unable to leverage valuable information from unlabelled images to help training. On the contrary, the results of our method indicate that three proposed strategies enable our method to effectively leverage the information of unlabelled images to assist in training on a limited quantity of labelled data and significantly boost performance. 
%fating
% \whsix{
% % We also report the distribution of top 8 importance score in testing set and the statistics of the pseudo labels generated by different methods in Figure \ref{fig:visualENCAA}. 
% We also report the statistics results on ENCAA in Figure \ref{fig:visualENCAA}. In particular,
% Figure \ref{fig:visualENCAA}(b) shows that in LP, PL and MT, most of unlabelled images' pseudo-labels are \ftfi{all} ``non-important'', resulting a bias to ``non-important'' class, \ie, the top-1 importance scores in most testing images are less than ours or even the softmax classifier threshold (0.5). In contrast, Figure \ref{fig:visualENCAA}(b) verifies that our method prevents the problem of classifying all individuals as ``non-important'' pseudo-labels and thus achieves more robust results; for instance, as shown in Figure \ref{fig:visualENCAA}(a), the gap between the importance scores of the most important people predicted by our method and other people's is larger than related methods.}

\begin{table}[t] 
	\centering
	{
		\resizebox{1\columnwidth}{!}{
		% \resizebox{\textwidth}{!}{
		\begin{tabular}{lccc|ccc}
			% \toprule
			
		  %  \#label   &  &  & 100 &  & & & & 300 & & \\
		  %  \midrule
		    % \toprule
		    Dataset  & \multicolumn{3}{c}{EMS}  & \multicolumn{3}{c}{ENCAA}  \\
		    % \midrule
		    \toprule
		    \#labelled images & $33$ \% & $66$ \% & $100$ \% & $33$ \% & $66$ \% & $100$ \%\\
			\midrule
			POINT (fully supervised) &$83.36$ &$85.97$ &$88.48$ & $84.60$ &$88.21$ &$89.75$ \\
			\midrule
		    \wh{Pseudo Label (PL)} &$83.37$ &$85.35$ &$88.57$ & $85.70$ &$88.43$ &$90.56$ \\
			\wh{Label Propagation (LP)} &$82.34$ & $86.33$&$86.66$ & $85.36$ &$88.61$ &$90.18$ \\
			\wh{Mean Teacher (MT)} &$84.50$ &$86.29$ &$87.55$ &$83.33$ &$84.66$ &$87.55$ \\
% 			\wh{Softmax} & $83.70$ & $86.81$ & $87.67$  \\
			\midrule
			{\bf Ours}&{$ \bf 87.81$}&{$\bf 88.44$} &{$\bf 89.79$} &{$\bf 88.75$} &{ $\bf 90.86$}&{$\bf 92.03$}\\
			% \midrule
			\bottomrule
% 			\vspace{-0.2cm}\\
% 			% \midrule
% 			% \toprule
% 			Dataset  & \multicolumn{3}{c}{ENCAA} \\
% 			\toprule
% 		    % \midrule
% 		    Percentage of labelled images &$33$ \% & $66$ \% & $100$ \% \\
% 			\midrule
% 			POINT (fully supervised) & $84.60$ &$88.21$ &$89.75$ \\
% 			\midrule
% 		    \wh{Pseudo Label (PL)} & $85.70$ &$88.43$ &$90.56$ \\
% 			\wh{Label Propagation (LP)} & $85.36$ &$88.61$ &$90.18$ \\
% 			\wh{Mean Teacher (MT)}  &$83.33$ &$84.66$ &$87.55$ \\
% % 			\wh{Softmax (SM)} &$84.35$ & $87.66$ & $89.93$ \\
% 			\midrule
% 			{\bf Ours}&{$\bf 88.75$} &{ $\bf 90.86$}&{$\bf 92.03$}\\
% 			\bottomrule
		\end{tabular}}%
		\vspace{0.1cm}
		\caption{Comparison with \whthi{related} methods on both datasets.}
		\label{tab:Comp}
	}
	\vspace{-0.3cm}
\end{table}%

%fating
% \begin{figure}[t]
% 	\begin{center}
% 		%		\fbox{\rule{0pt}{2in}\rule{0.9\linewidth}{0pt}}
% 		\includegraphics[width=0.95\linewidth]{./figures/visualENCAA.pdf}
% 		%		\vspace{-0.2cm}
% 		\centering\small\caption{\ftth{Fgiure (a) is the distribution of top 8 importance score in testing set in ENCAA datasets, Figure (b) is the statistics of unlabelled data’s pseudo-labels on ENCAA dataset. Better view in color.}
% 		}
% 		\label{fig:visualENCAA}
% 	\end{center}
% 	\vspace{-0.5cm}
% \end{figure}

\subsection{\whthi{Effect of the Proportion of Labelled Images}}

To further understand the factors that affect the performance of semi-supervised important people detection, we evaluate our method using different portions of labelled images. We randomly select $33$ \%, $66$ \% and $100$ \% of labelled images, and the remaining labelled images together with unlabelled images are used \wh{WITHOUT} labels. We report the results in \ftth{Table \ref{tab:Comp}, Table \ref{tab:Abla} and Table \ref{tab:OursLPMT}}. It is clearly observed that using more labelled data can boost the overall performance of important people detection, which also enables the semi-supervised model to estimate more accurate pseudo-labels for unlabelled images and further boost performance. It also indicates that developing a semi-supervised model that can correctly predict pseudo-labels and combine them with the labelled training set is necessary. \wh{From another point of view, the results shown in Table \ref{tab:Abla} imply that our method can consistently outperform the fully supervised approach \wh{as well as} related baselines and clearly \wh{demonstrate} the consistent efficacy of the three proposed strategies.}

\subsection{\ftsec{Ablation Study}}

\whfo{We conduct ablation study to investigate the effect of three proposed strategies (\ie, ranking-based sampling (RankS), importance score weighting \ftsec{(ISW)} and effectiveness weighting \ftsec{(EW)}) on important people detection and shown the results in Table \ref{tab:Abla}, where ``$\textup{Ours}_{\textup{w/o ISW and EW}}$'' indicates our method using RankS only.}

\begin{table}[t] 
	\centering
	% \resizebox{15cm}
    % \resizebox{}{15cm}
	{
	    \resizebox{1.0\columnwidth}{!}{
		\begin{tabular}{lccc|ccc}
			% \toprule
			
		  %  \#label   &  &  & 100 &  & & & & 300 & & \\
		  %  \midrule
		    % \toprule
		    Dataset  & \multicolumn{3}{c}{EMS} & \multicolumn{3}{c}{ENCAA} \\
		    \toprule
		    % \midrule
		    \#labelled images & $33$ \% & $66$ \% & $100$ \% & $33$ \% & $66$ \% & $100$ \%  \\
			% \midrule
			% POINT (fully supervised) &$83.36$ &$85.97$ &$88.48$ \\
			\midrule
			$\textup{Ours}_{\textup{w/o Ranks, ISW and EW}}$ & $83.70$ & $86.81$ & $87.67$ &$84.35$ & $87.66$ & $89.93$\\
			$\textup{Ours}_{\textup{w/o ISW and EW}}$ &$85.55$&$87.25$ &$88.53$ & $87.13$&$90.53$&\ftfi{$91.49$} \\
			$\textup{Ours}_{\textup{w/o EW}}$ &$86.34$& $87.45$&$89.67$ &$87.68$ &$90.60$&$92.00$ \\
			Ours &{$ 87.81$}&{$ 88.44$} &{$ 89.79$} &{$ 88.75$} &{ $ 90.86$}&{$ 92.03$}\\
			% \midrule
			% \midrule
			\bottomrule
		
		\end{tabular}}%
		\vspace{0.1cm}
		\caption{ \whthi{Ablation study on both datasets. RankS represents ranking-based sampling \whthi{while} ISW and EW indicate importance score weighting and effectiveness weighting, respectively. $\textup{Ours}_{\textup{w/o ISW and EW}}$ means our model \ftfi{without using} ISW and EW.}}
		\label{tab:Abla}
	}
	\vspace{-0.5cm}
\end{table}%

\whthi{In Table \ref{tab:Abla}, it is evident that \wh{all strategies can improve performance in most label regimes, and} the ranking-based sampling strategy attains the greatest improvement; for instance, on the ENCAA dataset, if 33 \% of labelled images are used, the method \ftth{``$\textup{Ours}_{\textup{w/o ISW and EW}}$''} outperforms the \ftth{``$\textup{Ours}_{\textup{w/o RankS, ISW and EW}}$''} by $2.78$ \%. 
This result clearly shows that the ranking-based sampling enables that the relatively high score should be labelled as ``important'' and the rest remain ``non-important'' when predicting pseudo-labels within each unlabelled image, 
preventing assigning all ``non-important'' or all ``important'' pseudo-labels during label guessing in an image.} 
\whfif{This is also verified by Figure \ref{fig:pseudo-labelling}, where our method correctly predicts pseudo-labels for all individuals (Figure \ref{fig:pseudo-labelling}(b)) during training \whsix{and estimate accurate importance scores at the end (\eg, Figure \ref{fig:pseudo-labelling}(a))} while current semi-supervised learning approaches (\ie, LP and MT) assign all individuals as ``non-important'' samples}

\whthi{From Table \ref{tab:Abla}, we also observe that adding importance score weighting (ISW) can \whfo{consistently albeit slightly} boost the performance (\eg, the results of \ftth{``$\textup{Ours}_{\textup{w/o EW}}$''} vs. those of \ftth{``$\textup{Ours}_{\textup{w/o ISW and EW}}$''} are \ftfi{$89.67$} \% vs. \ftfi{$88.53$ \%} on the EMS if all labelled images are used). This indicates that ISW is able to alleviate the problem of data imbalance and ultimately benefits the training of important people detection.} 

\begin{figure}[t]
	\begin{center}
		%		\fbox{\rule{0pt}{2in}\rule{0.9\linewidth}{0pt}}
		\includegraphics[width=0.9\linewidth]{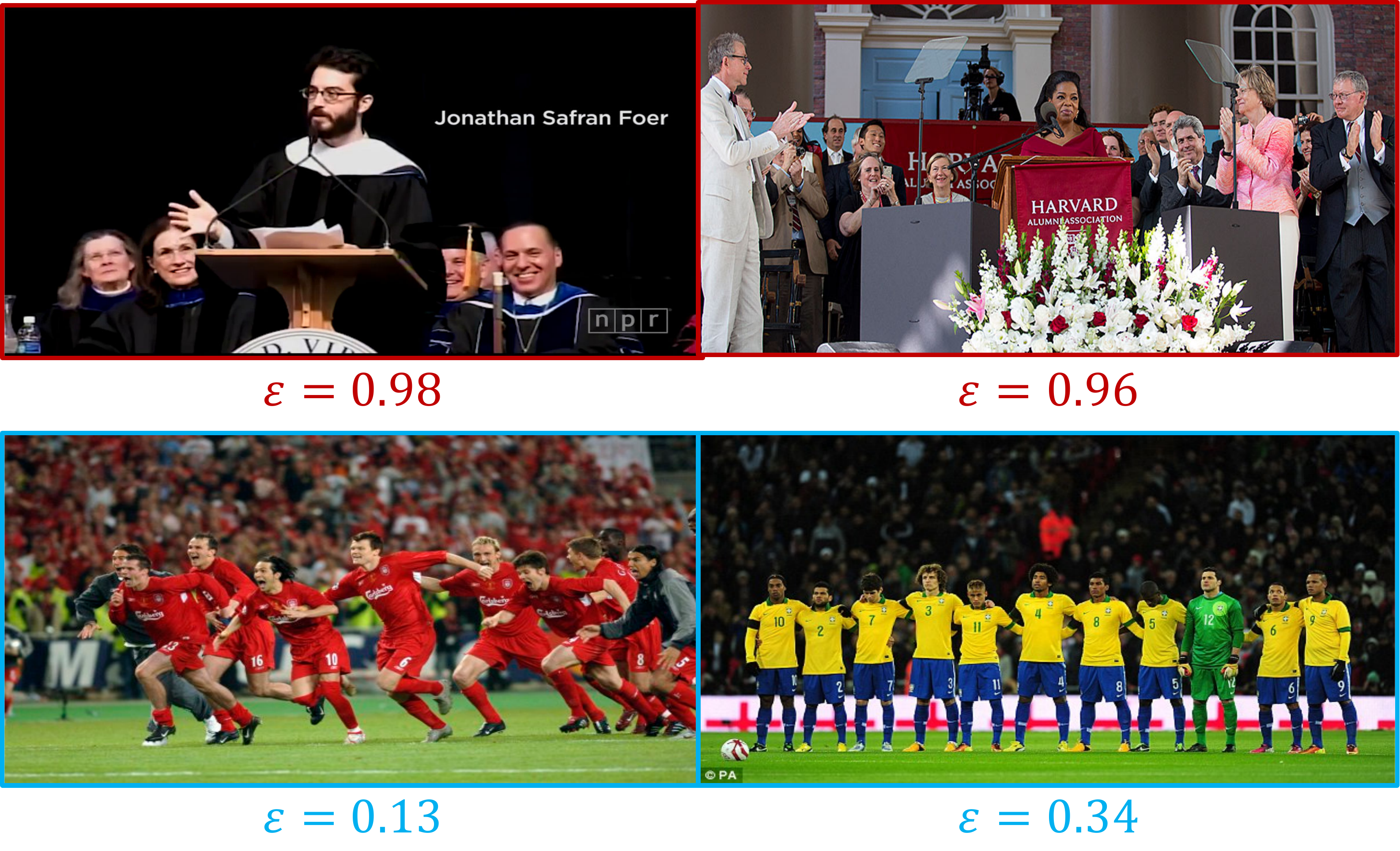}
		%		\vspace{-0.2cm}
		\centering\small\caption{Examples of unlabelled images and their effectiveness weights estimated automatically by our method.
		}
		\label{fig:w-example}
	\end{center}
	\vspace{-0.6cm}
\end{figure}

\whthi{In addition, comparing the full model %that uses three proposed strategies
and our model using both RankS and ISW, we clearly observe that the \wh{estimated} effectiveness weight (EW, defined in Eq. \ref{eq:EW}) improves the performance (\eg, \ftth{``Ours''} improves the performance of \ftth{``$\textup{Ours}_{\textup{w/o EW}}$''} from $86.34$ \% to $87.81$ \% on EMS if 33 \% of labelled images are used). This implies that our effectiveness weighting strategy is able to detect and neglect noisy unlabelled images with no important people, and this benefits important people detection. To further better understand how the effectiveness weight works, we visualize EW of several unlabelled images and present \wh{them} in Figure \ref{fig:w-example}. We clearly observe that if there are no important people in the unlabelled image, EW is small (as shown in the \wh{second row in Figure \ref{fig:w-example}}), while if important people are present, EW is nearly $1$ (as shown in the \wh{first row in Figure \ref{fig:w-example}}). This result again clearly demonstrates the efficacy of our proposed EW on detecting noisy images and \wh{neglecting} noisy samples during training.}

\begin{table}[t] 
	\centering
	% \resizebox{15cm}
    % \resizebox{}{15cm}
	{
	    \resizebox{1\columnwidth}{!}{
		\begin{tabular}{lccc|ccc}
			% \toprule
			
		  %  \#label   &  &  & 100 &  & & & & 300 & & \\
		  %  \midrule
		    % \toprule
		    Dataset  & \multicolumn{3}{c}{EMS}  & \multicolumn{3}{c}{ENCAA} \\
		    \toprule
		    % \midrule
		    \#labelled images & $33$ \% & $66$ \% & $100$ \% & $33$ \% & $66$ \% & $100$ \% \\
			% \midrule
			% POINT (fully supervised) &$83.36$ &$85.97$ &$88.48$ \\
% 			\mid\textbf{rule
% 			\wh{Label Propagation (LP)} &$82.34$ & $86.33$&$86.66$ \\
% 			\wh{M}ean Teacher (MT)} &$84.50$ &$86.29$ &$87.55$ \\
			\midrule
			$\textup{Ours}_{\textup{LP}}$ &{$ 87.51$}&{$ 88.10$}& {$ 89.65$} & {$ 88.95$} & {$ 91.06$}& {$ 91.98$}\\
			$\textup{Ours}_{\textup{MT}}$ &{$ 87.23$}& {$ 88.56$}&{$ 90.72$} & {$ 88.97$}&{$ 90.93$} & {$ 91.62$}\\
			Ours &{$ 87.81$}&{$ 88.44$} &{$ 89.79$} &{$ 88.75$} &{ $ 90.86$}&{$ 92.03$}\\
			% \midrule
			% \midrule
			\bottomrule
		
		\end{tabular}}%
		\vspace{0.1cm}
		\caption{
		\whfo{Evaluation of different techniques (\ie, LP and MT) when used for instantiating pseudo-label estimation function \ftfi{(\whsix{\ie,} $g(\cdot)$)} instead of using Softmax function.}
		}
		\label{tab:OursLPMT}
	}
	\vspace{-0.3cm}
\end{table}%

\ftfi{Additionally, we also evaluate the effect of different techniques (\ie, LP and MT) used to estimate importance score in our method during pseudo-labelling in Table \ref{tab:OursLPMT}, where ``$\textup{Ours}_{\textup{LP}}$'' implys our method using \ftfi{Label Propagation} technique for importance score estimation during pseudo-labelling.} 
It is clearly shown in Table \ref{tab:OursLPMT} that the variants of method using different techniques for importance score estimation yield similar results, which demonstrates the stableness of our methods.

\section{Conclusion}\label{sec:con}

% !TEX root = main.tex
In this work, we study semi-supervised learning in the context of important people detection and propose a semi-supervised learning method for this task. Compared with recent semi-supervised learning approaches, our method is shown to be able to effectively leverage the information of unlabelled images to assist in model training. We also conduct extensive experiments on important people detection by a semi-supervised learning method, and the results confirm that 1) the pseudo-labels of individuals in a given unlabelled image should \whsec{have the special pattern in important people detection (\ie, the relatively high score should be labelled as ``important'' and the rest remains ``non-important'')}, and our proposed ranking-based sampling is able to achieve this; 2) our importance score weighting can alleviate the imbalance problem and boost performance; and 3) enabling the model to neglect the noisy unlabelled images with no important people is important during semi-supervised learning. By our learning, we are able to avoid costly labelling on important people detection and achieve satisfactory performance. 

\section{Acknowledgement}

This work was supported partially by the National Key Research and Development Program of China (2018YFB1004903), NSFC(U1911401,U1811461), Guangdong Province Science and Technology Innovation Leading Talents (2016TX03X157), Guangdong NSF Project (No. 2018B030312002), Guangzhou Research Project (201902010037), and Research Projects of Zhejiang Lab (No. 2019KD0AB03).

{\small
\bibliographystyle{ieee_fullname}
\bibliography{refs}
}

\newpage
% \appendix
% \noindent\textbf{\LARGE{Appendix}}
\onecolumn
\section*{\LARGE{Appendix}}\label{s:appendix}
\appendix
% !TEX root = main.tex
% we may need a supplymentary material to report all results and some statistics of the datasets.
\section{\whfo{Details of both EMS and ENCAA datasets}}

\whfif{We present some labelled and unlabelled examples of both our proposed datasets (\ie, EMS and ENCAA) in Figure \ref{fig:DSexampleEMS} and Figure \ref{fig:DSexampleENCAA}.}
% \ftth{To better illustrate two augmented datasets (\ie, EMS and ENCAA), we also present some examples in Figure \ref{fig:DSexampleEMS} and Figure \ref{fig:DSexampleENCAA}}

\begin{figure}[h]
	% \vspace{-0.2cm}
	\begin{center}
		\label{fig:DSexampleEMS}
		%		\fbox{\rule{0pt}{2in}\rule{0.9\linewidth}{0pt}}
		\includegraphics[width=1.0 \linewidth ]{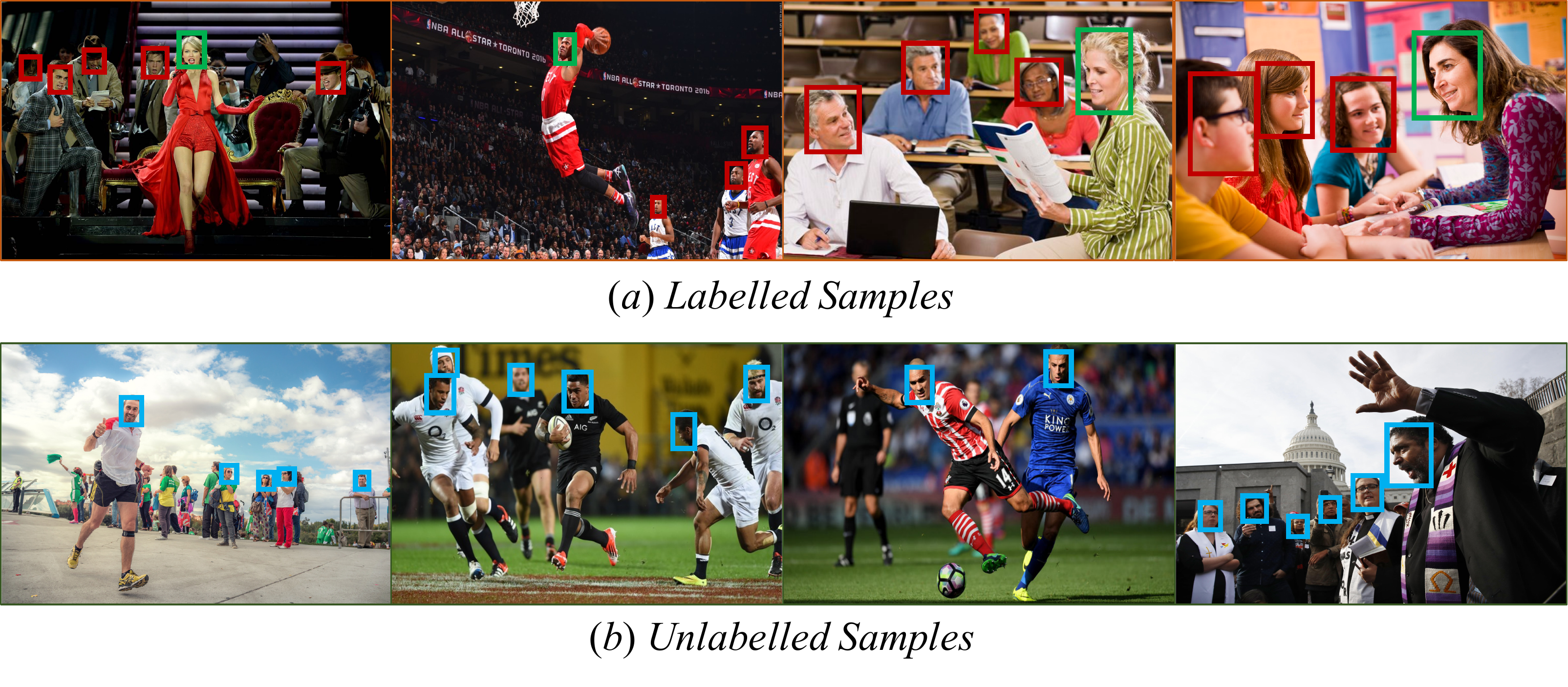}
		%		\vspace{-0.2cm}
		\centering\small\caption{\ftth{Examples of EMS Dataset. People are detected by face detectors, where in labelled images, people marked with green face bounding boxes are annotated as important people and people marked by red face bounding boxes are non-important people. In unlabelled images, all people are also detected by the face detectors \cite{li2019dsfd} (blue face bounding boxes).}}
		\label{fig:DSexampleEMS}
	\end{center}
	\vspace{-0.2cm}
\end{figure}

\begin{figure}[h]
	\begin{center}
		\label{fig:DSexampleENCAA}
		%		\fbox{\rule{0pt}{2in}\rule{0.9\linewidth}{0pt}}
		\includegraphics[width=1.0 \linewidth ]{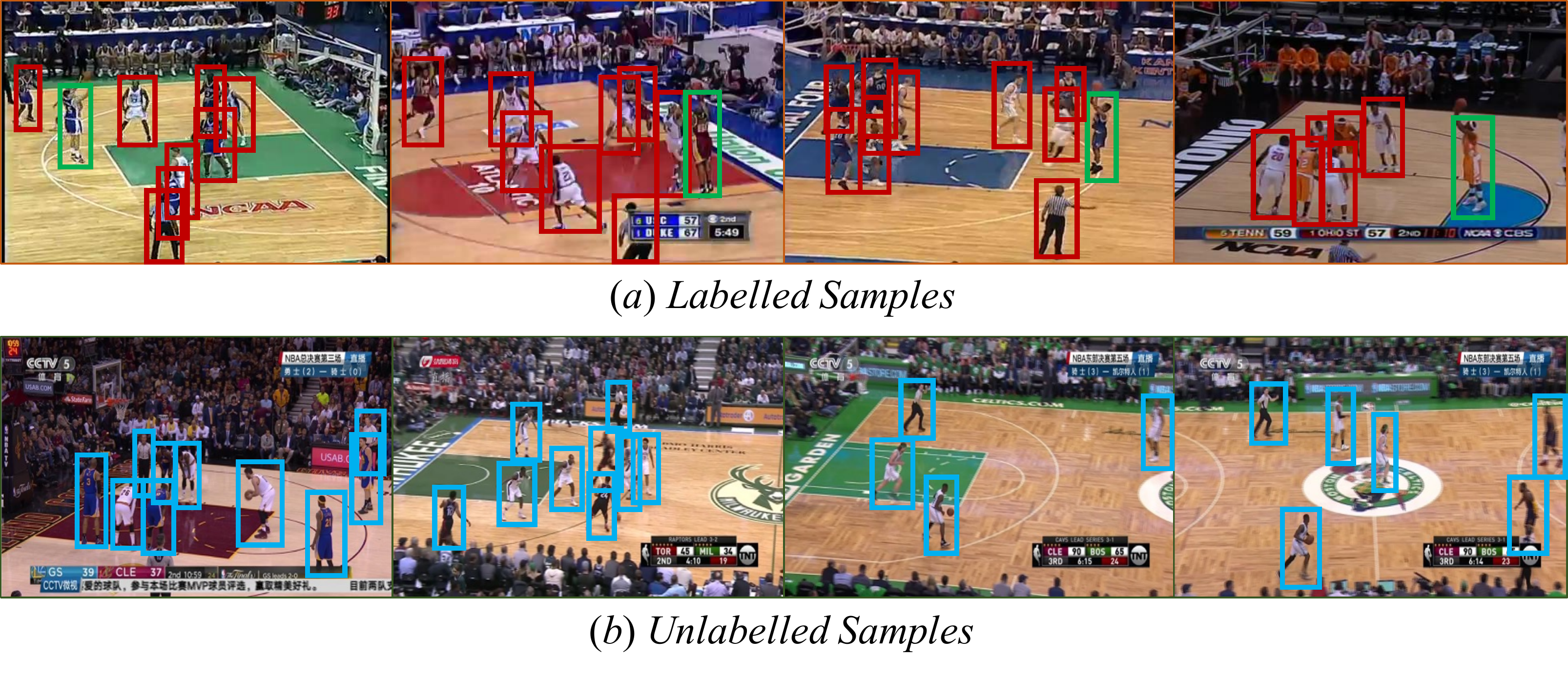}
		%		\vspace{-0.2cm}
		\centering\small\caption{\ftth{Examples of ENCAA Dataset. People are detected by object detectors, where in labelled images, people marked with green body bounding boxes are annotated as important people and people marked by red body bounding boxes are non-important people. In unlabelled images, all people are also detected by the object detectors \cite{redmon2018yolov3} (blue body bounding boxes).}}
		\label{fig:DSexampleENCAA}
	\end{center}{}
	\vspace{-0.4cm}
\end{figure}

% \section{Introduction}
\section{\whfo{Details of the basic Model}}\label{s:point}
% \subsection{Essential Knowledge Revisit}
In this work, we adopt the {\bf deep imPOrtance relatIon NeTwork (POINT)} \cite{Li_2019_CVPR} as the basic model for our method. More specifically, POINT
% The {\bf deep imPOrtance relatIon NeTwork (POINT)} \cite{Li_2019_CVPR} 
is an end-to-end network that automatically learns the relations among individuals in an image to encourage the network to formulate a more effective feature for important people detection. In particular, POINT has three components: 1) the \ftth{{\bf Feature Representation Module} containing ResNet-50, additional convolutional layers as well as fc layers, takes as input the social event images and detected individuals to \whfif{encode} features;} 2) \whfif{Fed with the encoded features, the }{\bf Relation Module} learns to model the relations graph among individuals and encode relation features from the graph; 3) taking as input the relation features encoded by the relation module, the {\bf Classification Module} \ftsec{together with a Softmax operator} transforms the relation feature into two scalar values indicating the probability of the ``important'' and ``non-important'' category.

\section{\whfo{\whfif{Additional} Results.}}

\begin{figure}[t]
	\begin{center}
		\label{fig:visualEMS}
		%		\fbox{\rule{0pt}{2in}\rule{0.9\linewidth}{0pt}}
		\includegraphics[width=1.0 \linewidth ]{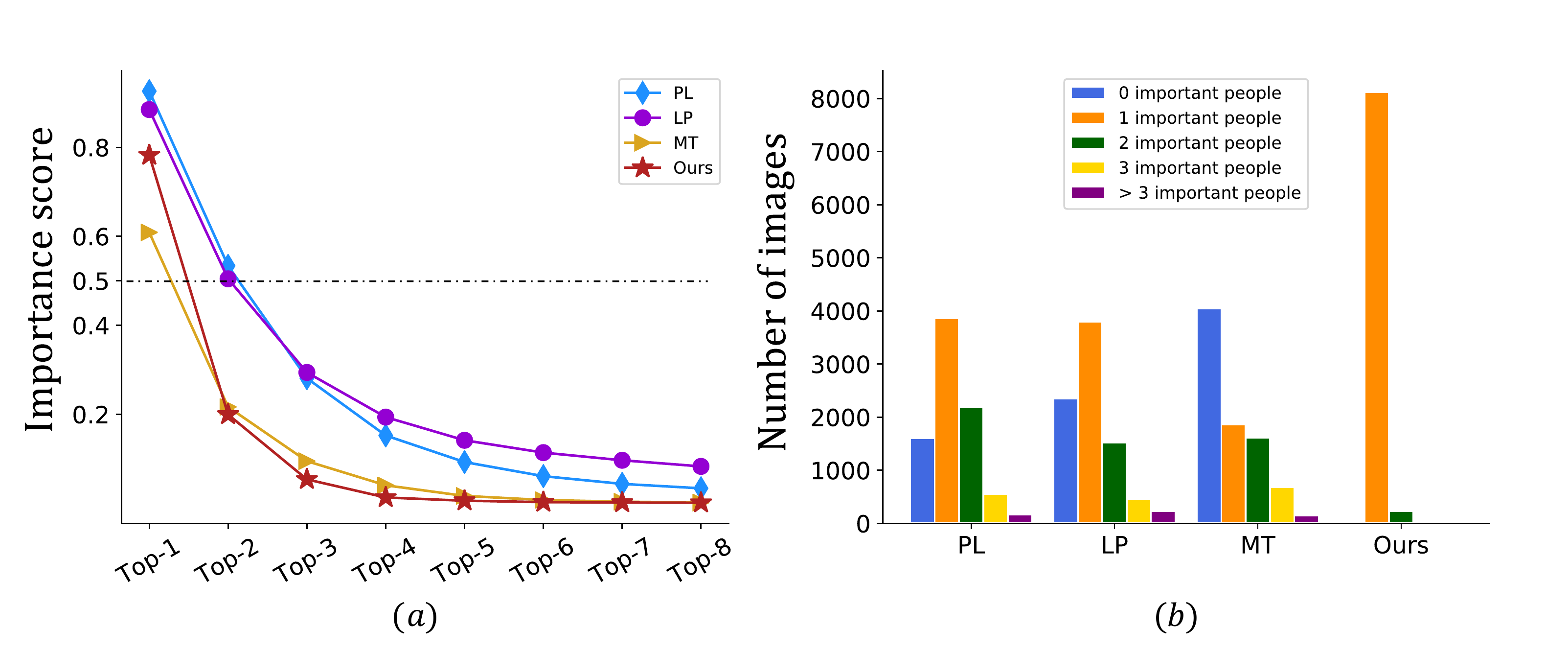}
		%		\vspace{-0.2cm}
		\centering\small\caption{\ftth{Fgiure (a) is the distribution of top 8 importance score in testing set in EMS datasets \whfif{and} Figure (b) is the statistics of unlabelled data’s pseudo-labels on EMS dataset. Better view in color.}
		% \ftsec{The distribution of important score for Top 8 people on test set in EMS for different methods.}
		}   
		\label{fig:visualEMS}
	\end{center}
	\vspace{-0.5cm}
\end{figure}
\begin{figure}[t]
	\begin{center}
		%		\fbox{\rule{0pt}{2in}\rule{0.9\linewidth}{0pt}}
		\includegraphics[width=0.95\linewidth]{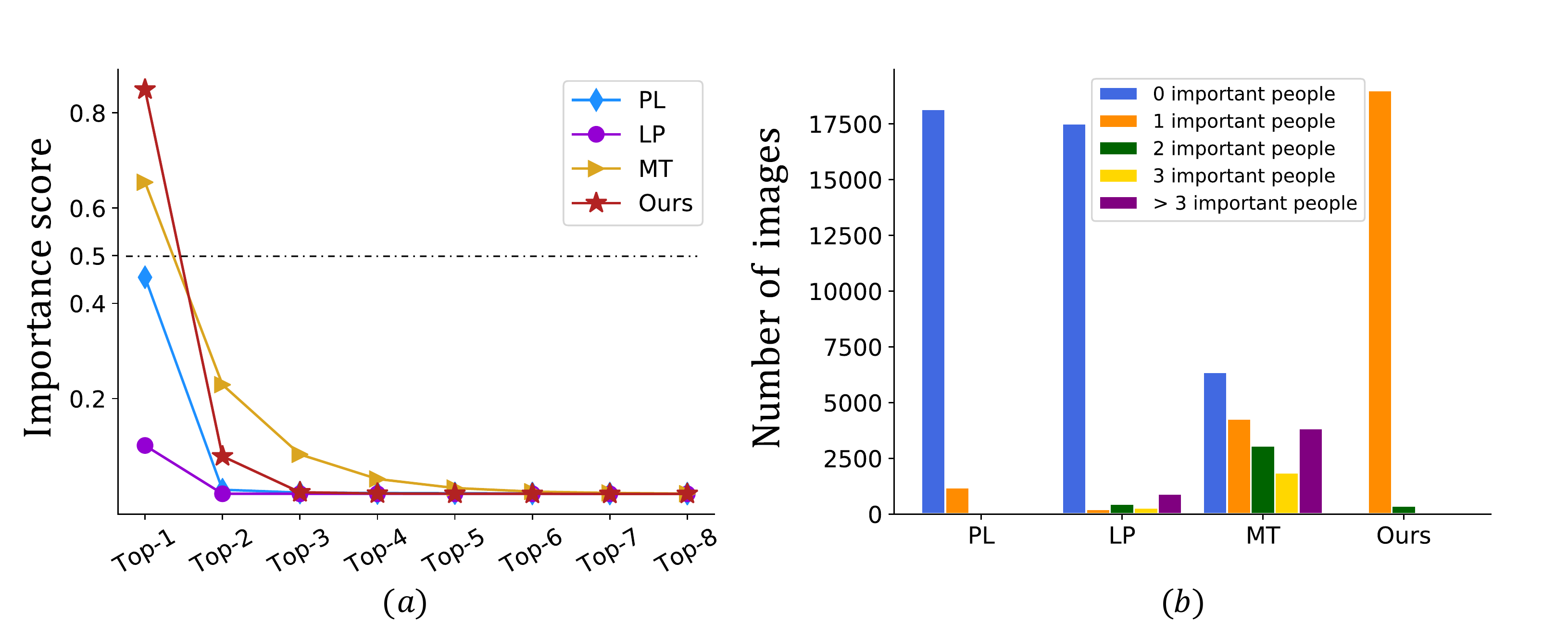}
		%		\vspace{-0.2cm}
		\centering\small\caption{\ftth{Fgiure (a) is the distribution of top 8 importance score in testing set in ENCAA datasets, Figure (b) is the statistics of unlabelled data’s pseudo-labels on ENCAA dataset. Better view in color.}
		}
		\label{fig:visualENCAA}
	\end{center}
	\vspace{-0.5cm}
\end{figure}
\subsection{\whfif{Statistics and Results on Testing Set \ftth{of EMS}.}}
To further compare our method with related methods, we illustrate the the statistics of unlabelled images' pseudo-labels \whfif{and the} distributions of importance score on testing set of ENCAA and EMS. 

We report the statistics results on ENCAA in Figure \ref{fig:visualENCAA}. In particular, Figure \ref{fig:visualENCAA}(b) shows that in LP, PL and MT, most of unlabelled images' pseudo-labels are \ftfi{all} ``non-important'', resulting a bias to ``non-important'' class, \ie, the top-1 importance scores in most testing images are less than ours or even the softmax classifier threshold (0.5). In contrast, Figure \ref{fig:visualENCAA}(b) verifies that our method prevents the problem of classifying all individuals as ``non-important'' pseudo-labels and thus achieves more robust results; for instance, as shown in Figure \ref{fig:visualENCAA}(a), the gap between the importance scores of the most important people predicted by our method and other people's is larger than related methods.
	
For EMS dataset, we calculate the number of important people in an unlabeled image according to their respective pseudo labels generated by different semi-supervised learning methods and report the statistics results in Figure \ref{fig:visualEMS}(b). \whfif{Additionally}, we report the average distribution of top 8 importance scores in Figure \ref{fig:visualEMS}(a). Note that, if there are N people in a testing image and N is smaller than 8, we record top N importance score. From Figure \ref{fig:visualEMS}(b), it is clear that most of unlabeled images's pseudo labels generated by related semi-supervised learning methods are all non-important (\ie, the blue bar) \whfif{and some of unlabelled images have more than 3 important people pseudo-labels (\ie, the purple bar)}. This again clearly points out \whfif{the imbalance pseudo-labelling} problem (\ie, assigning pseudo-labels of all detected persons in an image as ``non-important'' or ``important'') in recent state-of-the-art semi-supervised learning methods for important people \whfif{detection}. 
It is also demonstrated that by using our \whfif{proposed method, our method yields} a more reliable pseudo-labels (\ie, Figure \ref{fig:visualEMS}(b)), resulting in better testing prediction. In particular, by using our proposed strategies, we prevent the problem of assigning all ``non-important'' or all ``important'' pseudo-labels to people in unlabelled images. By using our model, \whfif{on} the testing data of \whfif{the EMS dataset}, the average most important people's score is above 0.5, the classification threshold in softmax classifier, and is higher than that of MT; the rest of people's score is below 0.5, which are classfied into ``non-important''. \whfif{Though LP and PL can have high score for the most important people, they mis-classify the second important people whose score should be less than 0.5}. \whfif{Beyond this}, {we can see that} the gap between the importance score of most important people and other people is larger than the compared methods.
\begin{table}[t] 
	\centering

	{
	    \resizebox{\columnwidth}{!}{
		\begin{tabular}{lcccccc}
			
		    Dataset  & \multicolumn{3}{c}{EMS}&\multicolumn{3}{c}{ENCAA} \\
		    \toprule
		    % \midrule
		    Percentage of labelled images & $33$ \% & $66$ \% & $100$ \%& $33$ \% & $66$ \% & $100$ \%  \\
			\midrule
			POINT (fully supervised) &$83.36$ &$85.97$ &$88.48$ & $84.60$ &$88.21$ &$89.75$\\
			\midrule
			% \wh{Label Propagation (LP)} &$82.34$ & $86.33$&$85.11$ \\
			\wh{Label Propagation (LP)} &$82.34$ & $86.33$&$86.66$ &$85.36$ & $88.61$&$90.18$   \\
			$\textup{Ours}_{\textup{LP}}^{\textup{w/o ISW and EW}}$ &$86.68$ &$87.47$& $88.41$ &$88.50$ & $90.59$ &$91.51$  \\
			$\textup{Ours}_{\textup{LP}}^{\textup{w/o EW}}$ &$86.88$&$87.92$& $89.38$\ &$88.82$ & $90.98$& $91.92$\\
			$\textup{Ours}_{\textup{LP}}$&{$\bf 87.51$}&{$\bf 88.10$}& {$\bf 89.65$} &{$\bf 88.95$} & {$\bf 91.06$}& {$\bf 91.98$}\\
			\midrule
			\wh{Mean Teacher (MT)} &$84.50$ &$86.29$ &$87.55$ &$83.33$ &$84.66$ &$87.55$  \\
			$\textup{Ours}_{\textup{MT}}^{\textup{w/o ISW and EW}}$ &$86.11$&$87.77$ & $88.93$ &$88.38$ & $90.35$& $91.16$ \\
			$\textup{Ours}_{\textup{MT}}^{\textup{w/o EW}}$ &$86.59$&$88.29$ &$89.49$  & $88.95$ &$90.63$ & $91.37$\\
			$\textup{Ours}_{\textup{MT}}$ &{$\bf 87.23$}& {$\bf 88.56$}&{$\bf 90.72$} & {$\bf 88.97$}&{$\bf 90.93$} & {$\bf 91.62$}\\
			\bottomrule
		\end{tabular}}%
		\vspace{0.1cm}
		\caption{\whthi{Ablation study on both datasets. RankS represents ranking-based sampling \whthi{while} ISW and EW indicate importance score weight and effectiveness weight, respectively. $\textup{Ours}_{\textup{MT}}^{\textup{w/o ISW and EW}}$ means our model using Mean Teacher for importance score estimation during pseudo-labelling and \ftfi{without using} ISW and EW.}}
		\label{tab:allAbla}
	}
	% \vspace{-0.3cm}
\end{table}%

\subsection{\whfif{Additional Ablation Study Results.}}
We also report additional ablation study results where we adopt Label Propagation (LP) and Mean Teacher (MT) for importance score estimation in our method during pseudo-labelling. \ftth{Specially, ``$\textup{Ours}_{\textup{MT}}$'' implys our method using Mean Teacher for importance score estimation during pseudo-labelling.} \whfo{The results in Table \ref{tab:allAbla} demonstrate that our method incorporating with existing semi-supervised learning is able to improve the performance over the one of the respective method, which strongly verifies that our method is generic and effective.
% our proposed strategies is able to be incorporated to existing semi-supervised learning approaches such that they can be applied for important people detection, verifying that our method is generic and effective. 
It is also clear that three proposed strategies can consistently improves the performance regardless of which baseline is built on. These results strongly imply the effectiveness and stableness of all proposed strategies.}

\subsection{\whfif{CMC Curves.}}
\begin{figure}[h]
	\begin{center}
		%		\fbox{\rule{0pt}{2in}\rule{0.9\linewidth}{0pt}}
		\includegraphics[width=1.0 \linewidth ]{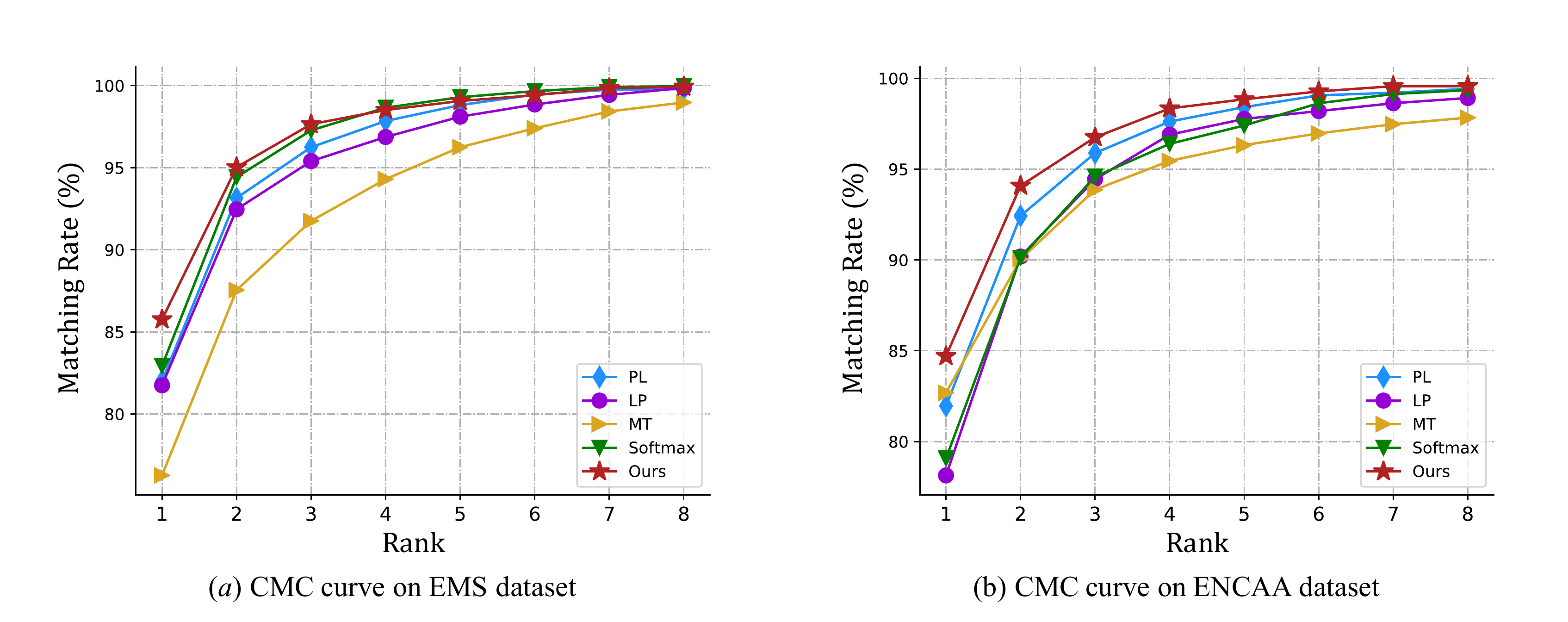}
		%		\vspace{-0.2cm}
		\centering\small\caption{ \ftsec{CMC curve on the EMS dataset (Figure (a))ENCAA dataset (Figure (b)).}
		% \ftsec{The distribution of important score for Top 8 people on test set in EMS for different methods.}
		}   
		\label{fig:cmc}
	\end{center}
	\vspace{-0.5cm}
\end{figure}

\ftsec{We \ftfif{also} plot the Cumulative Matching Characteristics (CMC) curves \cite{li2018personrank} of different methods on EMS and ENCAA datasets (Figure \ref{fig:cmc}. Compare with related semi-supervised methods, the results reported in the both figures show that ours method performs better for retrieving the important people from images, which indicates \whfif{that our method is able to leverage the information at unlabelled data to benefit important people detection.}}

\subsection{Visual Comparisons}
\begin{figure*}[h]
	\begin{center}
		%		\fbox{\rule{0pt}{2in}\rule{0.9\linewidth}{0pt}}
		\includegraphics[width=0.95 \linewidth ]{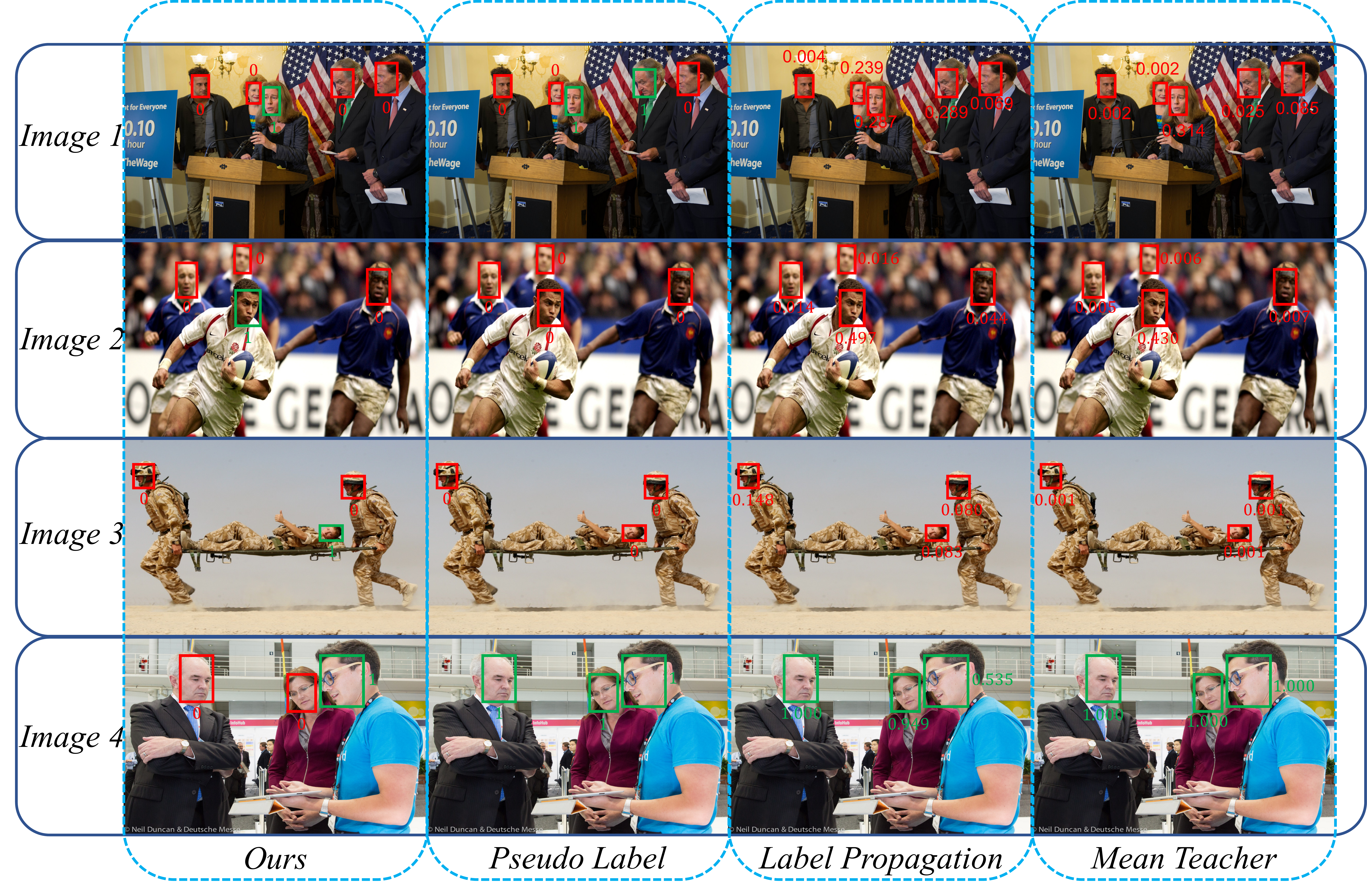}
		%		\vspace{-0.2cm}
		\centering\small\caption{ \ftth{Pseudo-labels of unlabelled images generated by our method and the compared methods (\ie Pseudo Label \cite{lee2013pseudo}, Label Propagation \cite{iscen2019label} and Mean Teacher \cite{tarvainen2017mean}) on EMS. The people with green bounding boxes are treated as ``important'' people, and people with red bounding boxes \whfif{are} \ftfif{classified as} ``non-important'' people  (the softmax classifier threshold is $0.5$) during pseudo-labelling.}
		% \ftsec{The distribution of important score for Top 8 people on test set in EMS for different methods.}
		}   
		\label{fig:pseudolabelEMS}
	\end{center}
	\vspace{-0.5cm}
\end{figure*}
\begin{figure*}[ht]
	\begin{center}
		%		\fbox{\rule{0pt}{2in}\rule{0.9\linewidth}{0pt}}
		\includegraphics[width=0.7 \linewidth ]{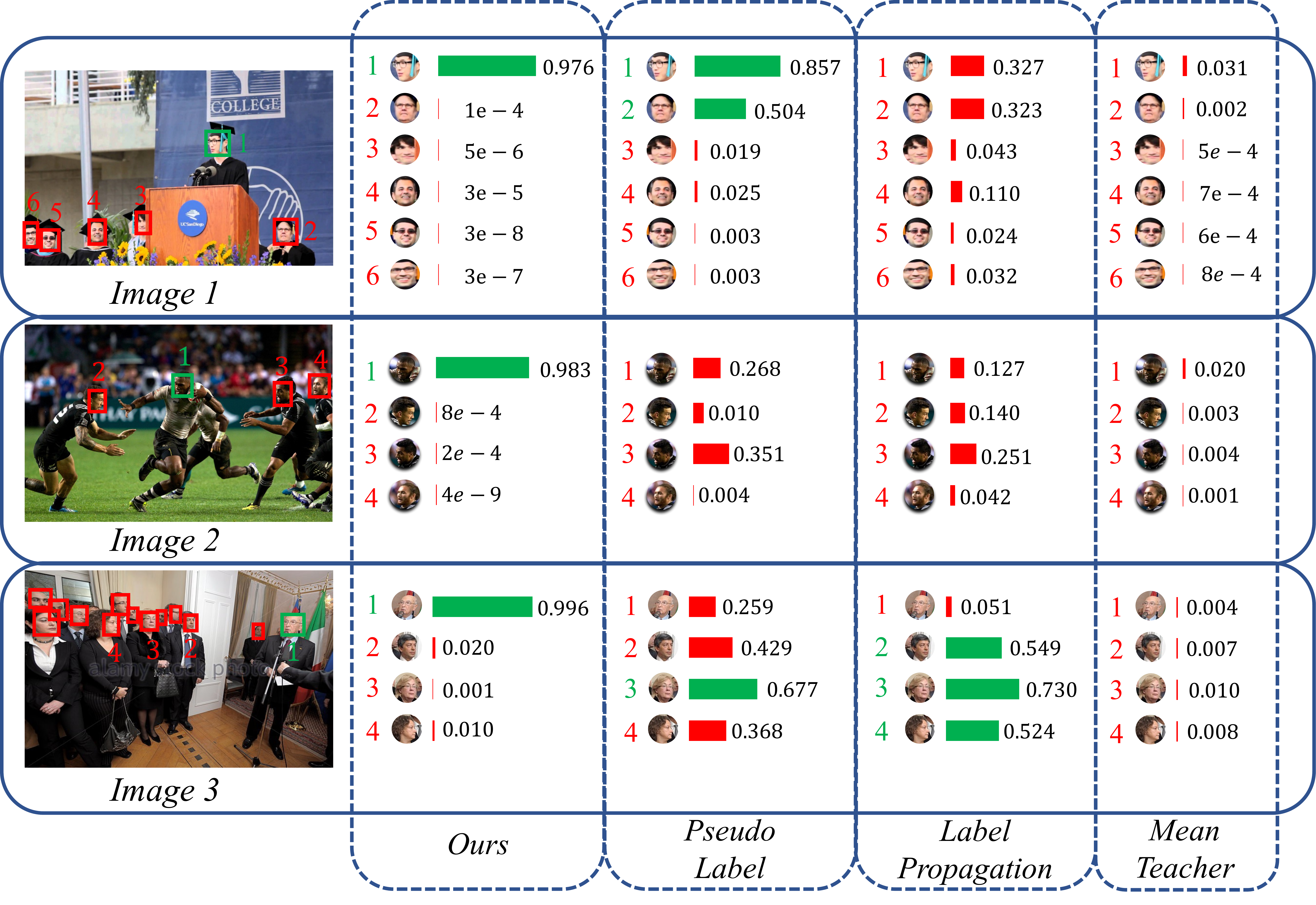}
		%		\vspace{-0.2cm}
		\centering\small\caption{ \ftth{The importance score \whfif{estimated by different methods on EMS. The images shown in the first column are testing images with ground-truth importance annotations, where people with green bounding boxes are the ground-truth ``important'' people and people with red bounding boxes are ``non-important'' people. In the second to fifth column, we report the importance score prediction of sampled people in respective testing images estimated by different methods.}}
		% \ftsec{The distribution of important score for Top 8 people on test set in EMS for different methods.}
		}   
		\label{fig:scoreEMS}
	\end{center}
	\vspace{-0.5cm}
\end{figure*}
% \cite{lee2013pseudo}, Mean Teacher (MT) \cite{tarvainen2017mean} and Label Propagation (LP) \cite{iscen2019label},
\begin{figure*}[h]
	\begin{center}
		%		\fbox{\rule{0pt}{2in}\rule{0.9\linewidth}{0pt}}
		\includegraphics[width=0.95 \linewidth ]{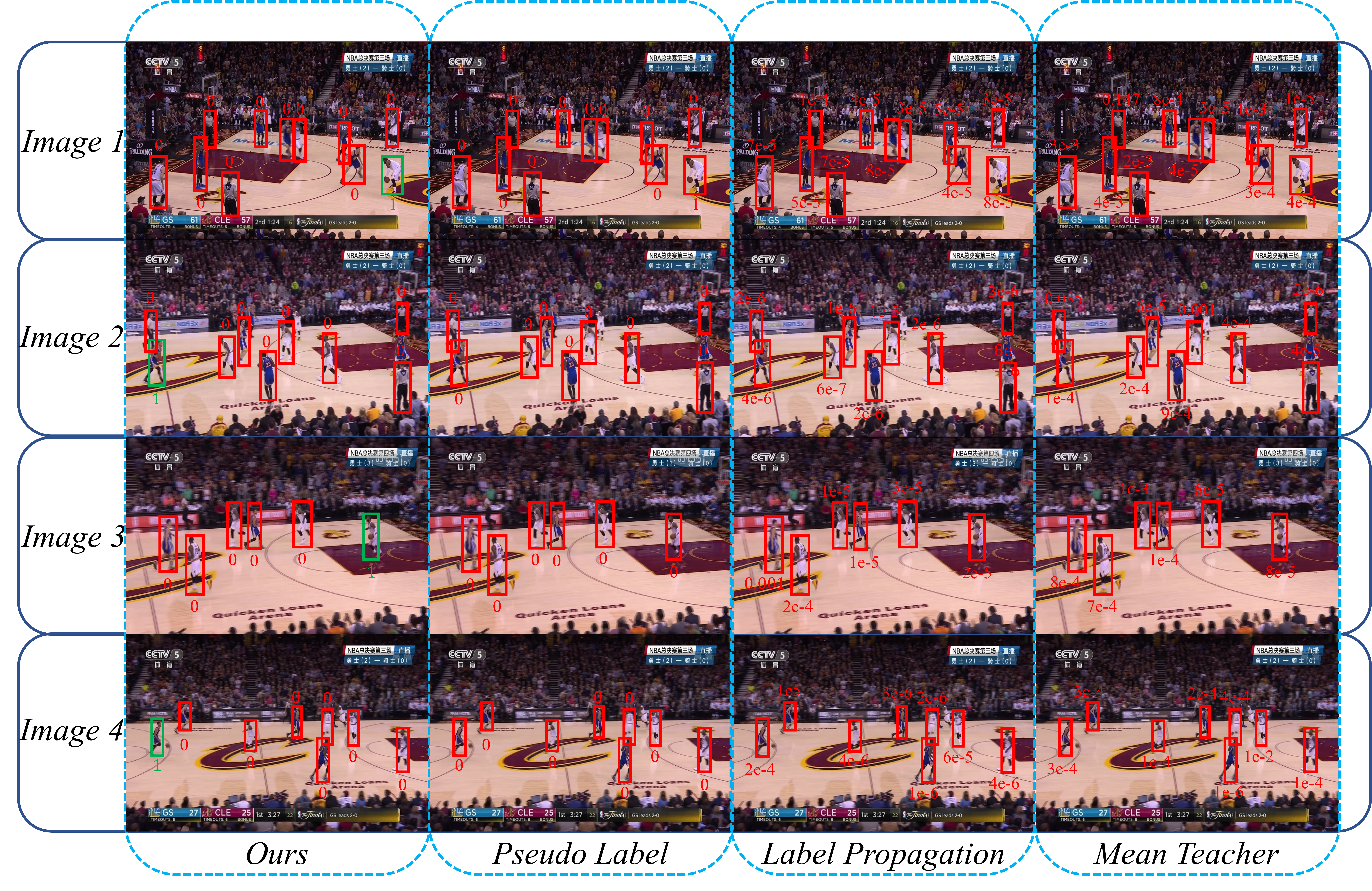}
		%		\vspace{-0.2cm}
		\centering\small\caption{ \ftth{Pseudo-labels of unlabelled images generated by our method and the compared methods on ENCAA. The people with green bounding boxes are treated as ``important'' people, and people with red bounding boxes \whfif{are} \ftfif{classified as} ``non-important'' people (the softmax classifier threshold is $0.5$) during pseudo-labelling.}
		% \ftsec{The distribution of important score for Top 8 people on test set in EMS for different methods.}
		}   
		\label{fig:pseudolabelENCAA}
	\end{center}
	\vspace{-0.5cm}
\end{figure*}
\noindent \textbf{\ftth{Pseudo-labels of unlabelled images generated by Different Methods.}}
\whfif{To better visualize the pseudo-labels generated by different methods, we present some examples of unlabelled images' pseudo-labels in Figure \ref{fig:pseudolabelEMS} and Figure \ref{fig:pseudolabelENCAA}. From both figures, it is worth noting that those recent semi-supervised would assign all ``non-important'' pseudo-labels to the people in an unlabelled image while our method is able to prevent this problem. Beyond this, we can clearly see that the pseudo-labels generated by our method are correct and this provides more reliable pseudo supervision for training.}
% \ftth{We also illustrate the pseudo-labels generated by our method and related methods in Figure \ref{fig:pseudolabelExample}. 
% It is worth noting that our method can estimate \ftfif{the most reasonable} pseudo-labels (the ``Ours'' column in Figure \ref{fig:pseudolabelExample}), while the related methods  assign all “non-important” or all “important” pseudo-labels to people in most cases. It certify that our pseudo-labelling is more helpful to the training process for important people detection than other methods.
% }

\begin{figure*}[ht]
	\begin{center}
		%		\fbox{\rule{0pt}{2in}\rule{0.9\linewidth}{0pt}}
		\includegraphics[width=0.7 \linewidth ]{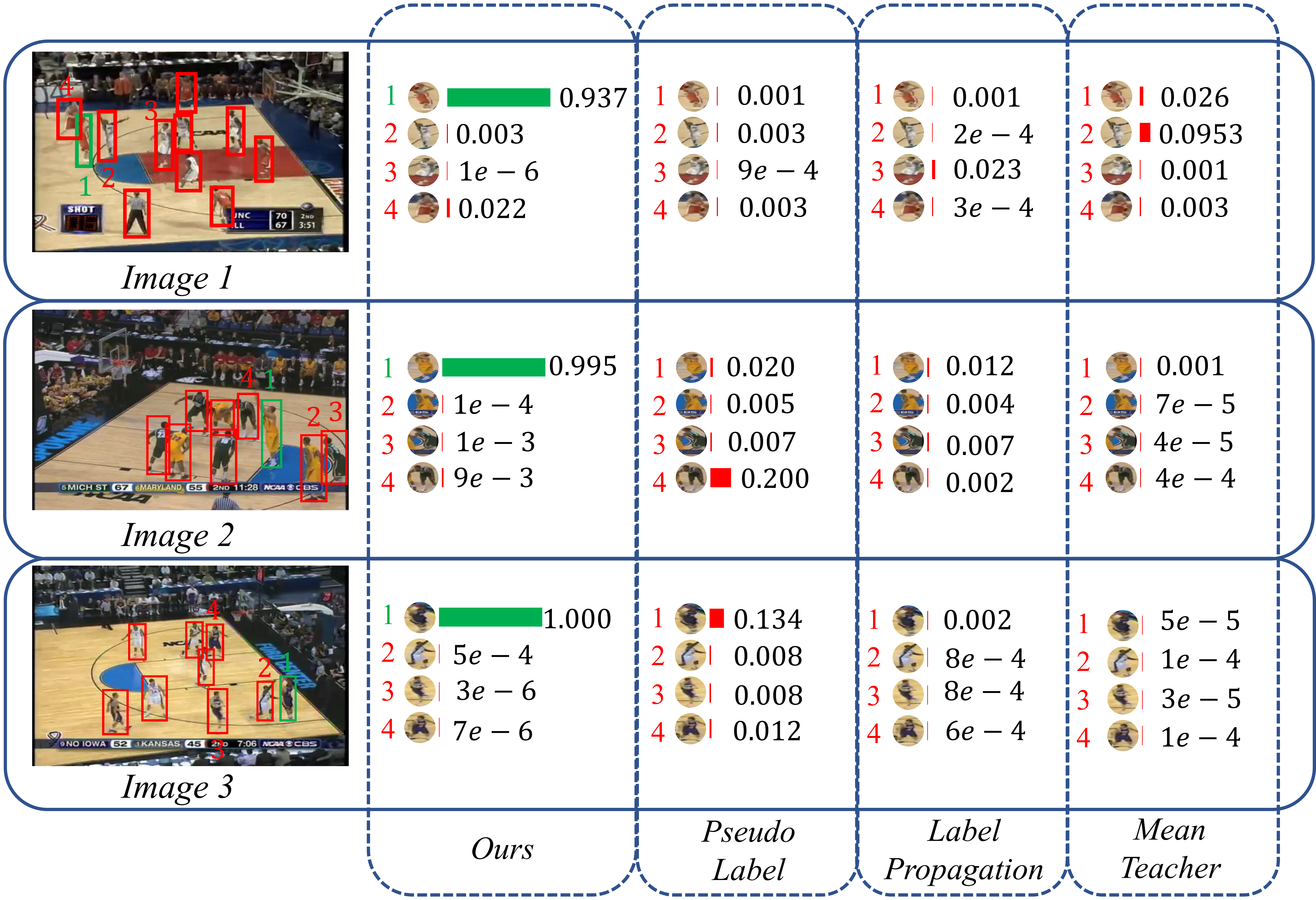}
		%		\vspace{-0.2cm}
		\centering\small\caption{ \ftth{The importance score \whfif{estimated by different methods on ENCAA. The images shown in the first column are testing images with ground-truth importance annotations, where people with green bounding boxes are the ground-truth ``important'' people and people with red bounding boxes are ``non-important'' people. In the second to fifth column, we report the importance score prediction of sampled people in respective testing images estimated by different methods.}}
		% \ftsec{The distribution of important score for Top 8 people on test set in EMS for different methods.}
		}   
		\label{fig:scoreENCAA}
	\end{center}
	\vspace{-0.5cm}
\end{figure*}
\noindent \textbf{\ftth{The Generated Importance Score of Different Methods.}}
\whfif{To better illustrate how our method affects the results, we report people in testing images' importance score estimated by different methods in Figure \ref{fig:scoreEMS} and Figure \ref{fig:scoreENCAA}. From two figures, it is clear that our method yields correct and very robust importance score prediction for testing images. In contrast, those compared methods are unable to identify the important people in images (\eg, Image 3 in Figure \ref{fig:scoreEMS}). Additionally, sometimes those compared method mis-classify all people as ``non-important'' category.}

\end{document}